%% file: arxiv.tex
\PassOptionsToPackage{dvipsnames}{xcolor}

\documentclass[10pt]{article} 
\usepackage[preprint]{tmlr}

\usepackage{easyReview}

\usepackage[utf8]{inputenc} 
\usepackage[T1]{fontenc}    

\usepackage{microtype}
\usepackage{graphicx}
\usepackage{wrapfig}
\usepackage{subcaption}
\usepackage{booktabs} 
\usepackage{etoc}
\usepackage[toc,page,header]{appendix}
\usepackage{minitoc}

\usepackage{hyperref}
\usepackage{soul}

\usepackage{amsmath}
\usepackage{amssymb}
\usepackage{mathtools}
\usepackage{amsthm}
\usepackage{bbold}
\usepackage{multirow}
\usepackage{bm}
\usepackage[capitalize,noabbrev]{cleveref}
\usepackage{authblk}
\usepackage{wrapfig,lipsum,booktabs}
\usepackage{tcolorbox}
\usepackage{todonotes}
\usepackage{color, colortbl}
\definecolor{Gray}{gray}{0.95}

\theoremstyle{plain}

\theoremstyle{definition}

\theoremstyle{remark}

\input{defs.tex}

\crefname{appendix}{App.}{Apps.}

\input{math_commands.tex}

\usepackage{hyperref}
\usepackage{url}

\title{Revisiting Knowledge Distillation: The Hidden Role of Dataset Size }

\author[1,2]{Giulia Lanzillotta}
\author[3]{Felix Sarnthein}
\author[2]{Gil Kur}
\author[2]{Thomas Hofmann}
\author[2]{Bobby He}
\affil[1]{ETH AI Center, Switzerland}
\affil[2]{Department of Computer Science, ETH Zurich, Switzerland}
\affil[3]{ELLIS Institute Tübingen, Germany}

\begin{document}
\doparttoc 
\faketableofcontents 

\maketitle
\begin{abstract}
The concept of knowledge distillation (KD) describes the training of a student model from a teacher model and is a widely adopted technique in deep learning.
However, it is still not clear how and why distillation works. Previous studies focus on two central aspects of distillation: \emph{model size}, and \emph{generalisation}. In this work we study distillation in a third dimension: {dataset size}.
We present a suite of experiments across a wide range of datasets, tasks and neural architectures, demonstrating that the effect of distillation is not only preserved but amplified in low-data regimes. We call this newly discovered property the \emph{data efficiency of distillation}. Equipped with this new perspective, we test the predictive power of existing theories of KD as we vary the dataset size. Our results disprove the hypothesis that distillation can be understood
as label smoothing, and provide further evidence in support of the dark knowledge
hypothesis. Finally, we analyse the impact of modelling factors such as the objective, scale and relative number of samples on the observed phenomenon.
Ultimately, this work reveals that the dataset size may be a fundamental
but overlooked variable in the mechanisms underpinning distillation.

\end{abstract}

\section{Introduction}

Knowledge distillation (KD) was introduced by \citet{bucilua2006model,hinton2015distilling} as a mechanism for \emph{transferring knowledge} between models with potentially different parameterizations. In its simplest form, the standard training targets are replaced by the soft predictions of a second model, referred to as the \emph{teacher}. Since its inception, KD has evolved into a widely adopted technique in deep learning, with numerous variants and applications across domains \citep{zagoruyko2016paying, passalis2018learning, park2019relational, tung2019similarity, Tian2020Contrastive, he2021feature, touvron2021deit, caron2021emerging, beyer2022knowledge}.

Research on KD has traditionally emphasized two aspects: model size and generalization. First, distillation enables a substantial reduction in model size without a corresponding drop in accuracy. Second, it can enhance generalization: a student model may outperform a teacher of identical architecture—a setting known as \emph{self-distillation}—even in the absence of additional supervision \citep{furlanello2018born}. Crucially, these results have typically been established under the assumption that teacher and student are trained on the same dataset.

\begin{figure*}[ht!]
\centering
{\includegraphics[width=0.6\textwidth]{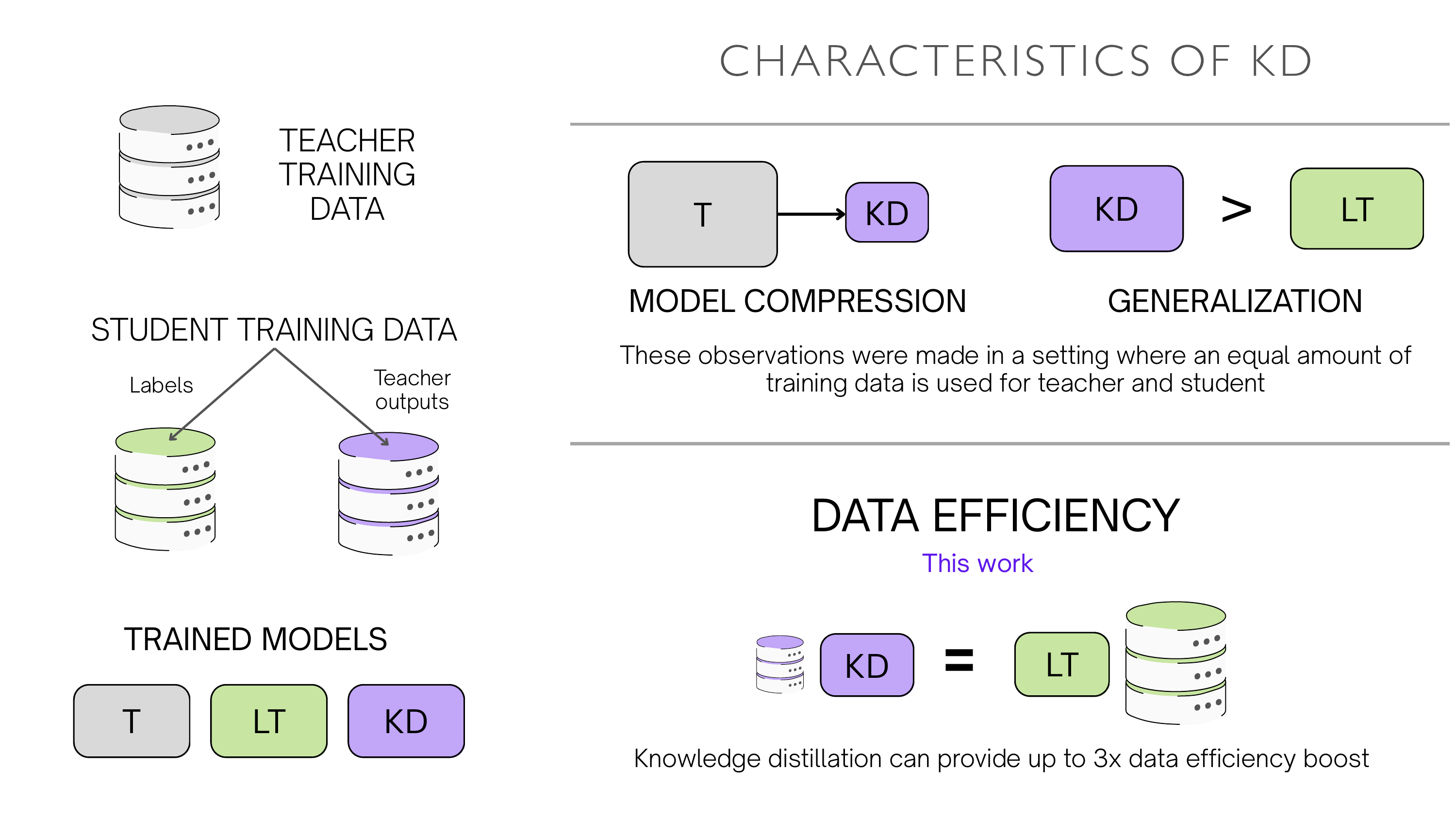}}
{\includegraphics[width=0.23\textwidth]{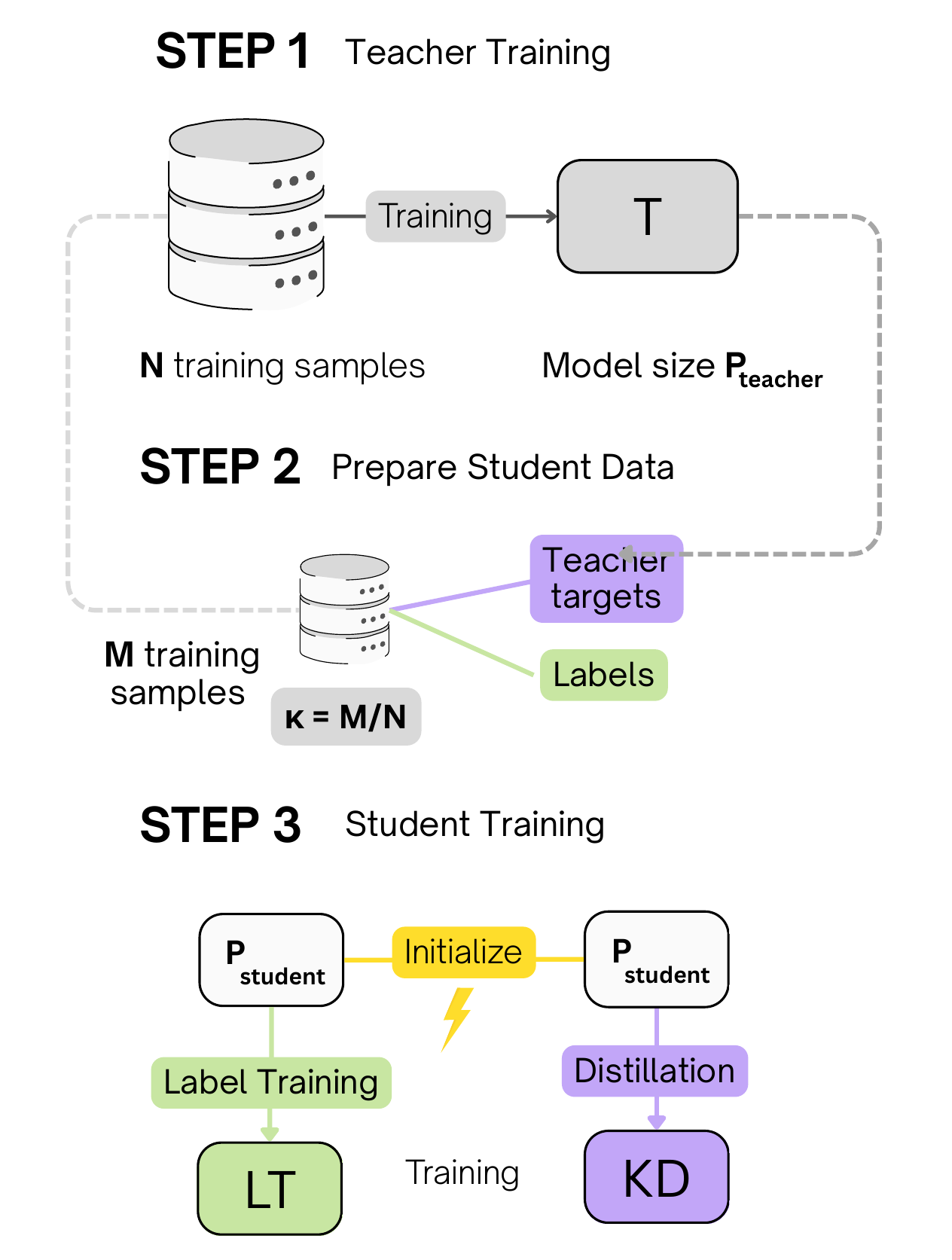}}
\caption{\textbf{Overview of our study.} (\textbf{Left and Center}) Schematic summary of the contributions of this work. Prior research has primarily emphasized two main benefits of knowledge distillation—model compression and improved generalization. We introduce a third, previously underexplored dimension: \emph{data efficiency}, showing that distillation yields the largest relative gains in low-data regimes. (\textbf{Right}) Experimental setup used throughout the paper. Students share the same architecture as the teacher or are smaller, and are trained with varying dataset fractions and temperatures. We systematically compare training with one-hot labels versus distillation targets to isolate the role of soft supervision across data scales.}
\label{figure1}
\end{figure*}

In this work, we examine KD through the lens of \emph{dataset size}, leading to the identification of a previously unreported property that we term the \emph{data efficiency} of distillation. \Cref{figure1} illustrates our setup and main finding. In brief, we observe that the performance advantage commonly attributed to self-distillation is amplified in low-data regimes and extends to heterogeneous teacher–student pairs trained on different amounts of data. More precisely, while prior work reported modest improvements of roughly $1\%$ in test accuracy under full-data training \citep{furlanello2018born, mirzadeh2020improved, mobahi2020self}, we find substantially larger relative gains—on the order of $10\%$—when using as little as $2\%$ of the data. Equivalently, with distillation, the same performance achieved with standard label supervision can be obtained using roughly three times less data. This effect is consistent across architectures (CNNs and Transformers) and modalities (vision and language); see \Cref{data-efficiency}.

These empirical findings resonate with theoretical results suggesting that KD improves statistical efficiency in fixed-feature settings (e.g., linear models or networks in the NTK regime) \citep{phuong2019towards, ji2020knowledge, panahi2022analysis, zhao2023towards, menon2021statistical}. To our knowledge, this work provides the first systematic empirical corroboration of these theoretical predictions on modern architectures and widely used benchmarks.

Despite its practical success, KD still lacks a unified theoretical account. Competing explanations have emphasized connections to label smoothing \citep{yuan2020revisiting, zhou2021rethinking}, fidelity to teacher predictions \citep{stanton2021does}, or enhanced feature learning \citep{allen2020towards, he2021feature}. Our results add a data-centric perspective that revisits these hypotheses. By exposing how distillation behaves across varying data regimes, we reveal empirical biases in current theories and provide new evidence for evaluating competing explanations.

In summary, this paper makes the following contributions:
\begin{itemize}
    \item \textbf{Characterizing the data efficiency of distillation beyond the full-data regime.} We systematically investigate how the benefits of distillation vary with dataset size, highlighting pronounced gains in low-data scenarios (\Cref{sec:part1}).
    
    \item \textbf{Evaluating existing theories of distillation.} We assess whether prevailing hypotheses—such as label smoothing, dark knowledge, and feature alignment—adequately explain the observed efficiency, identifying their limitations and the regimes in which they apply (\Cref{sec:part2}).
    
    \item \textbf{Quantifying the influence of modeling choices on distillation.} We analyze how factors such as temperature, target type (hard vs.\ soft), network scale, and dataset fraction shape the performance gains, providing a unified perspective on the determinants of distillation efficiency (\Cref{sec:part3}).
\end{itemize}

\section{Related Work}
\label{related-works}

We review prior work on knowledge distillation (KD) in two steps. First, we summarize the main theoretical narratives proposed to explain why distillation improves generalization. Second, we discuss existing references to data efficiency in KD. Overall, the literature has largely focused on generalization and knowledge transfer, while the role of dataset size remains comparatively underexplored. For comprehensive surveys, we refer readers to \citet{moslemiSurveyKnowledgeDistillation2024a, liuTheoreticalPerspectivesKnowledge2025a}.

\subsection{Theories of Distillation}
\paragraph{Dark knowledge.} 
A dominant explanation for the benefits of KD emphasizes the \emph{dark knowledge} embedded in a teacher’s predictive distribution \citep{hinton2015distilling}. According to this view, distillation is effective because soft predictions encode inter-class similarities that are absent from one-hot labels. Building on this idea, \citet{allen2020towards} proposed the \emph{multi-view feature hypothesis}, whereby students, owing to independent initializations, learn complementary features, and distillation succeeds by transferring features that the student would not otherwise discover.

\paragraph{Label smoothing.} 
A competing line of work interprets KD as a form of \emph{label-smoothing regularization} \citep{szegedy2016rethinking}. \citet{yuan2020revisiting, zhou2021rethinking} argue that the generalization improvements observed under KD largely stem from the implicit regularization induced by softened targets \citep{muller2019does}. Further evidence \citep{furlanello2018born, sarnthein2023random} suggests that dark knowledge alone cannot account for KD’s empirical advantages, particularly when the inter-class structure provided by the teacher is random.

\paragraph{Fidelity.} 
Another perspective questions the assumption that successful KD requires the student to closely match teacher predictions. \citet{furlanello2018born, stanton2021does, nagarajan2023student} show that \emph{fidelity}—the agreement between student and teacher on test examples—is often lower than expected, and that higher fidelity does not necessarily correlate with improved generalization. This challenges the notion that the student’s role is merely to replicate the teacher’s decision boundary.

Beyond these main threads, several other studies provide complementary theoretical and empirical insights into KD, including analyses of optimization dynamics and representational transfer \citep{mobahi2020self, lopez2015unifying, dong2019distillation, yim2017gift, beyer2022knowledge, zhao2022decoupled}.

\subsection{Data Efficiency in Distillation}
The idea that KD may improve data efficiency was already hinted at in the original work of \citet{hinton2015distilling}, who observed reduced overfitting in low-data settings. However, the discussion was brief and inconclusive. Subsequent formal analyses have explored this question more rigorously. \citet{phuong2019towards} studied KD in linear classification, demonstrating faster statistical convergence when training students on teacher predictions. This result was later extended to infinite-width neural networks in the NTK regime \citep{ji2020knowledge}, which similarly exhibit improved convergence guarantees. While theoretically insightful, these works rely on fixed-feature assumptions that neglect feature learning—a key factor in practical neural networks \citep{chizat2019lazy, yang2020feature, allen2020towards}. 

Other theoretical perspectives, such as the bias–variance analysis of \citet{menon2021statistical} and subsequent refinements \citep{foster2019hypothesis, panahi2022analysis, zhao2023towards}, also point to potential data-efficiency benefits. Yet these results often yield vacuous bounds or apply to idealized settings far removed from practical neural networks. 

Empirical studies of data efficiency in KD have become increasingly relevant with the advent of large models. The high cost of teacher queries has motivated research on reducing the number of samples required during distillation. For instance, \citet{hsieh2023distilling} proposed a modified KD objective that improves sample efficiency for language models. Other works tackle low-data regimes by selecting more informative distillation data \citep{heDAKDDifficultyAwareKnowledge}, refining the loss formulation \citep{xuComparativeKnowledgeDistillationa}, or optimizing teacher selection \citep{wuScalingLawDataEfficient2025}.

Despite these efforts, the mechanisms underlying data efficiency in KD remain poorly understood. Most existing studies exploit rather than explain this efficiency, and a systematic analysis across data regimes is still lacking. In contrast, our work isolates the effect of dataset size under a minimal and controlled setup—without altering the distillation objective—providing new insights into when and why KD confers data efficiency relative to direct label training.

\section{Notation and Setting}
\label{setup}

We consider a $K$-class classification problem. The goal is to estimate the conditional distribution $\mathcal{P}(Y \mid X)$ from a dataset $D = \{(x,y)\} \subseteq \mathcal{X} \times \mathcal{Y}$ using a neural network. Let $f_\theta : \mathcal{X} \to \mathbb{R}^K$ denote the network mapping, with $z = f_\theta(x) \in \mathbb{R}^K$ the corresponding \emph{logits}. For a temperature parameter $\tau > 0$, the \emph{softmax} function is defined as
\[
\sigma_\tau(z)_k = \frac{\exp(z_k/\tau)}{\sum_{j=1}^K \exp(z_j/\tau)} \quad \text{for } k=1,\dots,K.
\]
We denote the resulting predictive distribution by $p_f^\tau(x) = \sigma_\tau(f_\theta(x))$.
For two probability distributions $p$ and $q$ over the same domain, we write the (expected) Kullback–Leibler divergence as
\[
\mathrm{KL}(p \,\|\, q) = \mathbb{E}_{x \sim D}\!\left[p(x)^\top \log\,\frac{p(x)}{q(x)}\right].
\]

\paragraph{Distillation objective.} 
Let $p_t^\tau(x)$ denote the teacher’s output distribution at temperature $\tau$, and let $\delta(y)$ be the one-hot distribution of the true label. The student network is trained by minimizing a convex combination of two terms:
\begin{equation}
\label{eq:distillation-loss}
\mathcal{L}_{\alpha}(f) 
= (1-\alpha) \, \mathbb{E}_{x \sim D}\,\left[ \mathrm{KL}\,\big(p_t^\tau(x) \,\|\, p_f^\tau(x)\big) \right]
+ \alpha \, \mathbb{E}_{(x,y) \sim D}\,\left[ \mathrm{KL}\,\big(\delta(y) \,\|\, p_f(x)\big) \right].
\end{equation}
Here $\alpha \in [0,1]$ interpolates between the two extremes: when $\alpha=1$, the loss reduces to standard cross-entropy training on labels (\emph{label training}); when $\alpha=0$, it corresponds to pure distillation from the teacher.

\subsection{Experimental Setup.} 
We adopt a controlled setup to directly compare distillation and label training, illustrated in \cref{figure1}. Starting from a trained teacher, we train two students with identical architectures and hyperparameters: one using teacher logits (KD) and one using ground-truth labels (LT). This yields a \emph{student pair} $(p_{\mathrm{KD}}, p_{\mathrm{LT}})$ that differs only in the source of supervision.

For a teacher trained on $N$ samples and a student trained on $M$ samples, we define the relative dataset fraction as $\kappa := M / N$, and we repeat experiments across values of $\kappa \in (0, 1)$.
Distillation is implemented with a temperature parameter $\tau$ following \citet{hinton2015distilling}, tuned separately per dataset.
Experiments are conducted on both image classification and autoregressive language modeling tasks, representative of standard KD settings. A full description of datasets, architectures, and training protocols is provided in \cref{details}.
Evaluation metrics are \emph{test error} or \emph{test accuracy} $\mathrm{Acc}(\cdot)$ for vision tasks and \emph{test perplexity} $\mathrm{PPL}(\cdot)$ for text.

\section{Empirical study of the data efficiency of distillation}\label{sec:part1}

\begin{figure*}[ht!]
\begin{center}\centerline{\includegraphics[width=\textwidth]{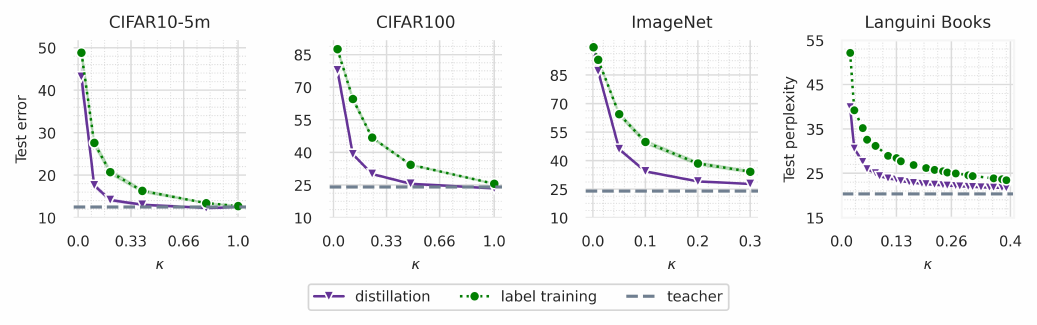}}
\caption{\textbf{Knowledge distillation improves sample efficiency.} Test error (for image classification) and perplexity (for autoregressive language modelling) as a function of the relative training dataset size $\kappa$, averaged over $5$ seeds. We compare models of the same architecture trained with either label training or knowledge distillation. Distillation consistently dominates label training in the low data regime.}
\label{data-efficiency}
\end{center}
\vskip -0.2in
\end{figure*}

We quantify the advantage of KD over LT by defining the \emph{performance increment} (PI):
\[
\mathrm{PI} := \mathrm{R}(p_{\mathrm{KD}}) - \mathrm{R}(p_{\mathrm{LT}}),
\]
where $\mathrm{R}$ denotes a generic performance metric---test accuracy for classification and negative test perplexity for language modeling.  
Alternatively, we consider the \emph{performance gain} (PG),
\[
\mathrm{PG} := \frac{\mathrm{R}(p_{\mathrm{KD}})}{\mathrm{R}(p_{\mathrm{LT}})},
\]
where $\mathrm{R}$ is defined analogously (using the inverse of perplexity for text).  
Both measures provide consistent interpretations: higher PI or PG indicates a greater advantage of distillation over label training. We adopt these definitions throughout the paper to compare the two training paradigms across data regimes.

In \cref{data-efficiency}, we report test performance as a function of the data fraction $\kappa$. Across all datasets, architectures, and modalities considered, distillation dominates label training whenever $\kappa < 1$. The gains are substantial: PI peaks between $0.05$ and $0.3$ in $\kappa$, reaching approximately $10\%$ on CIFAR10, $25\%$ on CIFAR100, $15\%$ on ImageNet, and $10\%$ on Languini Books. These increments far exceed those typically reported in self-distillation studies at $\kappa = 1$ (e.g., $0.20\%$ on CIFAR10 and $1.3\%$ on CIFAR100 in \citet{furlanello2018born}), demonstrating that the effect of distillation is not only preserved but amplified in low-data regimes.

Taken together, these results establish what we term the \emph{data efficiency of distillation}: when the training dataset available to the student is smaller than that of the teacher, distillation significantly boosts generalisation relative to direct label training. This effect is robust across modalities—vision and language—and across model scales. Beyond its empirical significance, the phenomenon raises a conceptual question: 
\begin{quote}
    \emph{Why should transferring soft labels from a teacher be most beneficial precisely when data is scarce?}
\end{quote}
We explore this question in \cref{sec:part2,sec:part3}. In the remainder of this section, we focus on the practical implications of this data efficiency and quantify the benefits students gain under different low-data scenarios. 

\subsection{Data efficiency versus computational efficiency}

\begin{wrapfigure}{r}{0.35\columnwidth}
    \vspace{-0.5in}
    \centering
    \includegraphics[width=\linewidth]{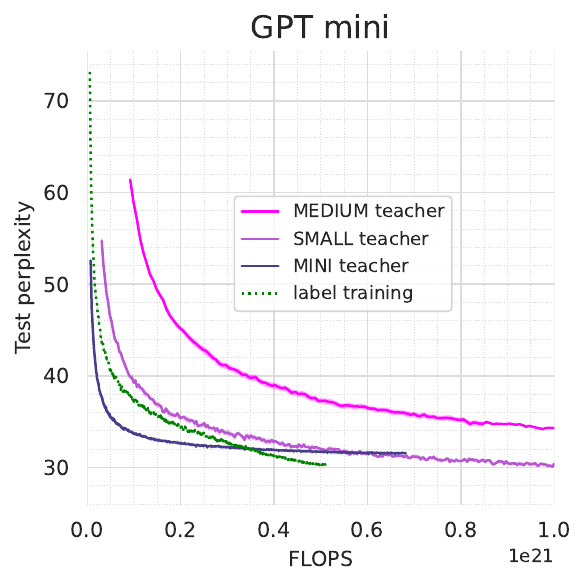}
    \caption{Test perplexity over FLOPs for GPT-MINI students trained with either distillation or label training, varying teacher size.}
    \label{fig:flops}
    \vspace{-0.5in}
\end{wrapfigure}

A natural question is whether the data efficiency of distillation also translates into reduced computational cost. To investigate this, we train GPT-MINI students on the Languini Books dataset using either knowledge distillation (KD) or standard label training, keeping the student architecture fixed. For KD, we consider three teachers of increasing size—GPT-MINI, GPT-SMALL, and GPT-MEDIUM—allowing us to study how teacher scale affects both final performance and training cost (see \cref{details} for model specifications).  

In \cref{fig:flops}, we report test perplexity as a function of total floating point operations (FLOPs), including both student training and the additional forward passes required for distillation. While larger teachers improve final performance, they also increase computational overhead. In two of the three teacher-student configurations, KD is actually less computationally efficient than label training, despite clear gains in data efficiency. These results demonstrate that the benefits of distillation in low-data regimes do not automatically yield computational savings, though they remain highly relevant for understanding the mechanisms that enhance generalization.

\subsection{Cross-data transfer via distillation}
\label{practice-DED}

We next examine whether the data efficiency of distillation can be leveraged in low-data transfer learning scenarios. We simulate such scenarios by using pre-trained teachers from the PyTorch hub, fine-tuning only their linear heads on the available dataset (\cref{transfer-table}). While prior work has studied KD under distribution shifts \citep{zhang2023revisiting}, our focus is specifically on settings with limited data. 

We evaluate several publicly available datasets varying in size and number of classes. Students are trained with either label training or distillation using all available data. Since the teacher has been pre-trained on ImageNet-21k, the effective relative dataset size $\kappa$ is small in nearly all cases. Hyperparameter tuning is critical in these low-data settings; to avoid biasing results in favor of distillation, \emph{we first optimize hyperparameters for label training and apply the same settings to distillation}. Hyperparameters are also tuned when training the teacher linear head.  

As shown in \cref{transfer-table}, distillation consistently outperforms label training across almost all tasks, with the exception of the largest dataset. This striking result indicates that KD can be highly beneficial in applications with severely constrained data, suggesting that further investigation of low-data distillation could unlock additional performance gains.  

\begin{table}[ht]
\centering
\vskip 0.1in
\begin{small}
\begin{sc}
\resizebox{\columnwidth}{!}{%
\begin{tabular}{l|cccccc}
\toprule
 & FLOWERS & DTD & AIRCRAFT  & CALTECH & CARS  & FOOD\\
\midrule
\rowcolor{Gray} \# Training samples & $1020$& $1880$ & $6667$  & $7810$ & $8144$  & $75750$\\
\rowcolor{Gray} \# classes & $102$ & $47$  & $10$  & $101$ & $196$ &  $101$\\
\midrule
Distillation &  $\bm{41.37} \pm 1.60$ & $\bm{37.04} \pm 6.01$ & $\bm{54.71} \pm 1.86$ & $\bm{73.98} \pm 0.82$ & $\bm{73.84} \pm 0.65$  &  $75.09 \pm 0.26$\\
Labels$^\star$ & $35.84 \pm 1.41$ & $28.36 \pm 2.45$   &  $ 53.40 \pm 4.30$ & $71.64 \pm 1.03$ & $70.20 \pm 1.54$ & $\bm{81.84} \pm 0.34$\\
\midrule
Teacher$^\star$& $\bm{86.60}$ & $\bm{67.44}$  &  $45.18$ & $\bm{94.00}$ & $55.03$  & $70.76$\\
\bottomrule
\end{tabular}%
}
\end{sc}
\end{small}
\vskip -0.1in
\caption{\textbf{Distillation with transfer learning.} Validation accuracy for distillation and label training on several datasets. Teacher is pretrained on ImageNet-21k and adapted by retraining only the linear head. Hyperparameter tuning ($^\star$) is applied for teacher linear head and student label training; the same settings are then used for distillation. Distillation outperforms label training in nearly all cases.}
\label{transfer-table}
\end{table}

\section{Existing Theories of Distillation}
\label{sec:part2}
Several hypotheses have been proposed to explain the mechanisms underlying knowledge distillation (KD), each supported by empirical evidence. In this section, we reproduce canonical experiments in our setup to evaluate which of these intuitions extend to the low-data or high-data regime ($\kappa \neq 1$) and which may be artifacts of prior studies limited to $\kappa = 1$. In particular, we investigate the roles of label smoothing, feature alignment (dark knowledge), and fidelity, with the goal of understanding their contribution to the data efficiency of distillation.

\subsection{Label Smoothing}

\newcolumntype{g}{>{\columncolor{Gray}}c}
\begin{table}[ht]
    \centering
    \vskip -0.1in
    \begin{tabular}{c|gcc|gcc}
        \toprule
                & \multicolumn{3}{c}{CIFAR100} &  \multicolumn{3}{c}{CIFAR10} \\
         \midrule
         $\kappa$ &  LT & LS & KD & LT & LS & KD \\
         \midrule
         0.02 & $12.44{\pm0.81}$ & $+0.48\pm 0.49$ & $+9.77 \pm 1.10$
            & $56.92\pm0.46$ & $-0.66\pm0.53$ & $+4.74\pm0.86$\\
         0.1 & $35.36\pm0.84$ & $+0.38\pm0.68$ & $\textbf{+25.46}\pm0.76$ 
           & $74.20\pm0.28$ & $-0.84\pm0.31$& $\textbf{+6.01}\pm0.33$\\
         0.2 & $53.21\pm0.44$ & $+0.48\pm0.68$ &$+16.72\pm0.53$
              & $78.82\pm0.47$ & $-0.22\pm0.44$& $+4.80\pm0.65$\\
         0.4 & $65.66\pm0.24$ &$\textbf{+0.60}\pm0.51$&$+8.75\pm0.54$
              & $82.35\pm0.28$ & $-0.16\pm0.38$ & $+3.26\pm0.30$\\
         1.0 & $74.42\pm0.22$ & $+0.47\pm0.41$ &$+2.12\pm0.24$
             & $85.43\pm0.15$& $\textbf{+0.28}\pm0.23$& $+1.53\pm0.24$\\
         \bottomrule
    \end{tabular}
    \vskip 0.1in
    \caption{\textbf{Distillation is data efficient, label smoothing is not.} Classification accuracy of label training (LT), and PI of label smoothing (LS) and knowledge distillation (KD) on CIFAR10 and CIFAR100. }
\label{label-smoothing-table}
\end{table}

A prominent line of work interprets KD as a form of \emph{label smoothing regularization} \citep{szegedy2016rethinking}, whereby the improved generalization observed with distillation is attributed to the softening of the target distribution rather than the inter-class information contained in the teacher's logits \citep{yuan2020revisiting, zhou2021rethinking, muller2019does}. To test this hypothesis  we replicate an experiment by \citet{yuan2020revisiting} in the context of low-data regimes. We compare distillation with a manually constructed label-smoothing baseline (LS) where the one-hot targets $\delta(y)$ are softened with a uniform probability mass on non-target classes (implementation details in \cref{hyperparameters}).

We report results across CIFAR10 and CIFAR100 in \cref{label-smoothing-table}. For each student, we compute the \textit{performance increment} (PI) relative to standard label training (LT).

While LS yields minor improvements over LT that remain roughly constant across $\kappa$, KD exhibits substantially higher PI in low-data regimes. Thus, although when using $100\%$ of the dataset label smoothing and distillation show similar PIs, their behaviour is substantially different for $\kappa<1$. This confirms that
the properties of distillation are not fully captured by label smoothing, which allows us to ultimately reject this hypothesis. 

\subsection{Dark knowledge}
\label{feature-learning-section}

\begin{figure}
\begin{center}
{\includegraphics[width=0.95\columnwidth]{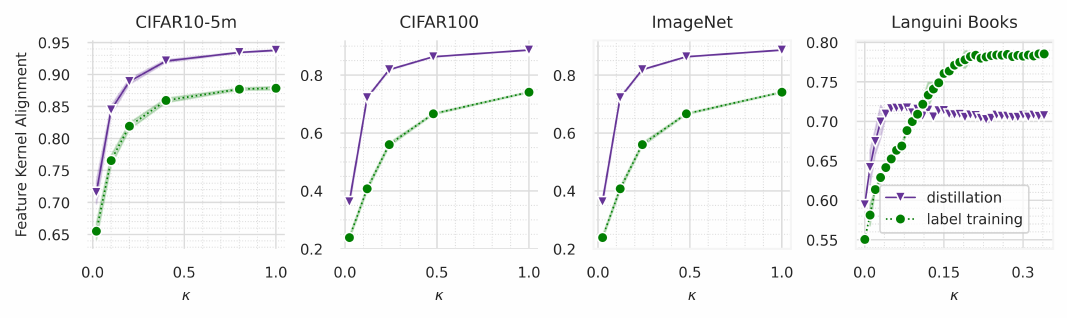}}
\caption{\textbf{Distillation induces feature kernel alignment in image classification settings.} On the y-axes the CKA of the feature kernels $k_\phi$ of the KD and LT students to the teacher's feature kernel. Note that the LT students and the LT teacher are both trained with labels. On the x-axis the portion of dataset used. We observe that KD produces markedly steeper curves, yielding high feature kernel alignments at low $\kappa$.  } 
\label{feature-kernel-alignment}
\end{center}
\vspace{-10pt}  
\end{figure}

\begin{figure}[ht]
\begin{center}
{\includegraphics[width=0.2\columnwidth]{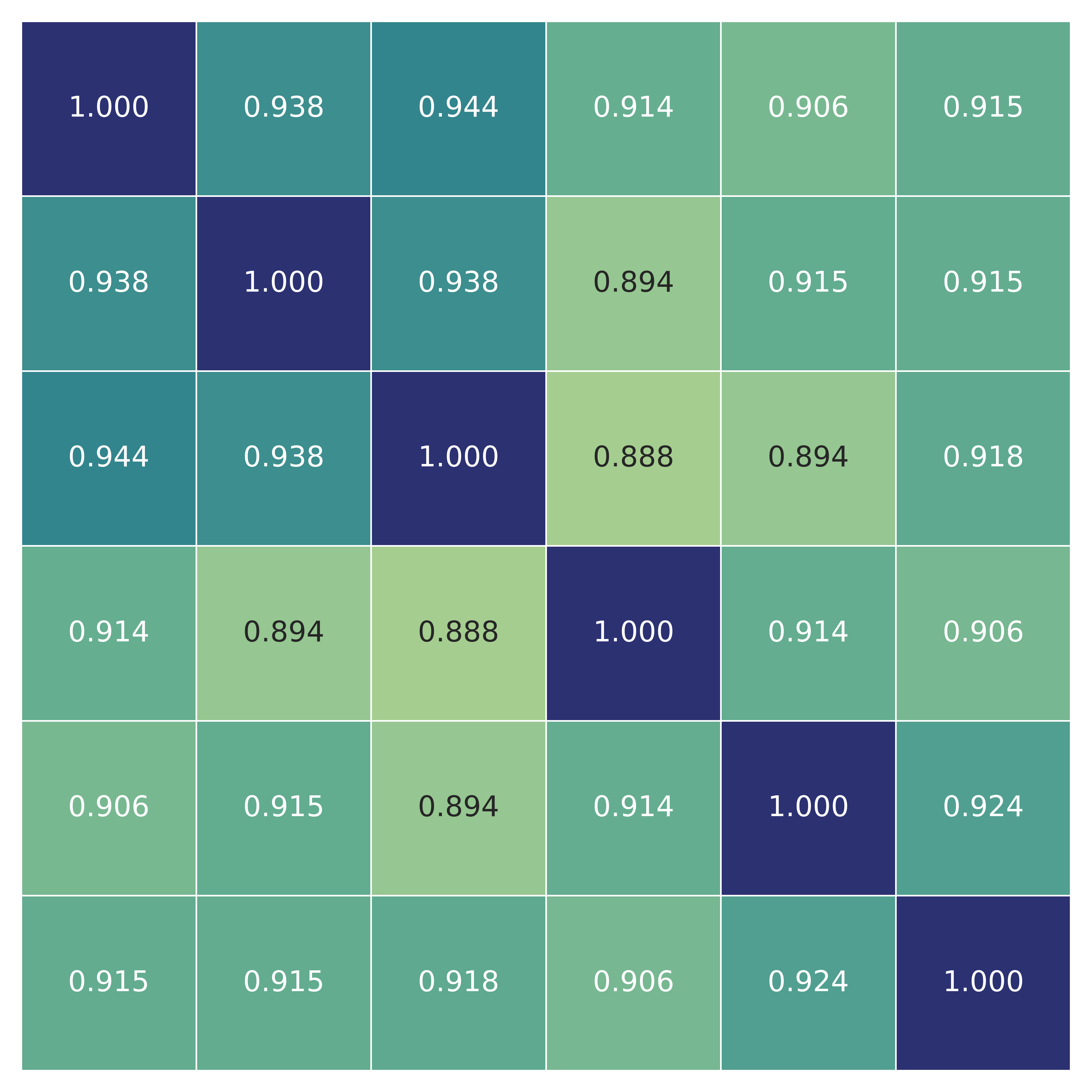}}
{\includegraphics[width=0.2\columnwidth]{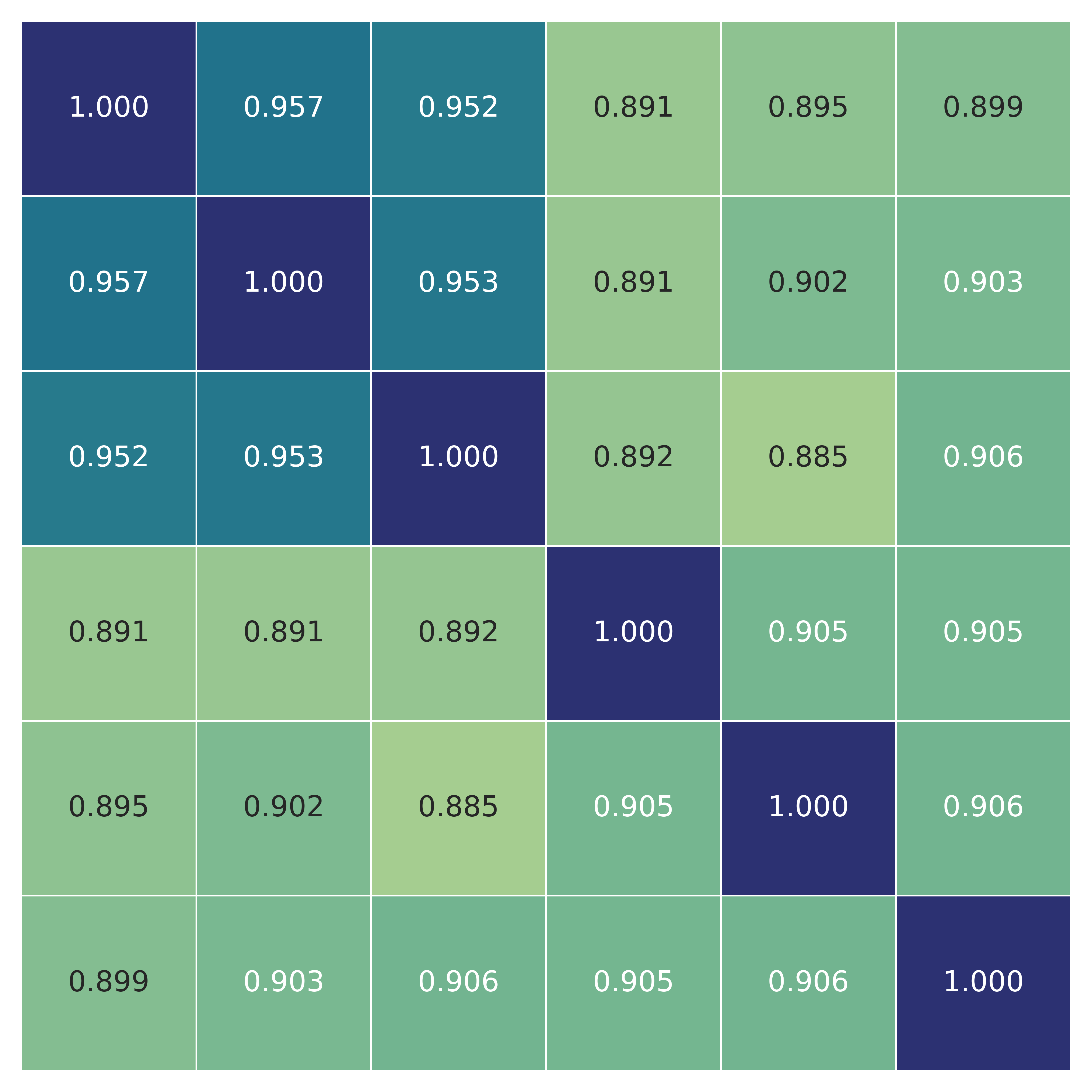}}
{\includegraphics[width=0.2\columnwidth]{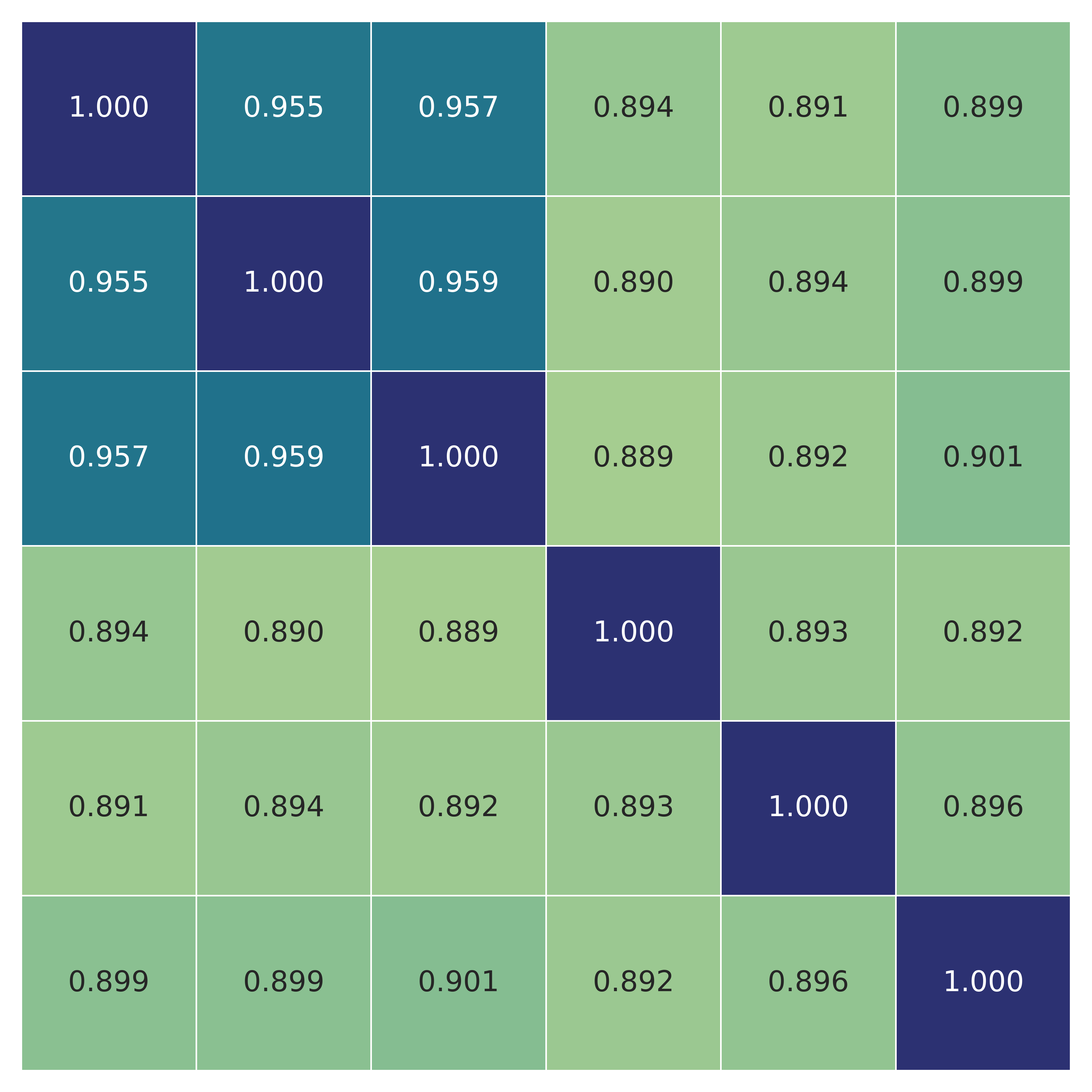}}
\caption{\textbf{Distilled students form a compact cluster.} Feature-kernel alignment between students trained with distillation or one-hot labels on the same fraction of CIFAR10 from three different initializations (six networks total). Entries correspond to pairwise alignments; the first three rows and columns represent distilled students, the last three label-trained students. From left to right: $\kappa=0.02$, $\kappa=0.1$, and $\kappa=0.2$. Temperature $\tau=20$ in all plots.}
\label{cross-cka}
\end{center}
\end{figure}

Another hypothesis posits that KD transfers \emph{dark knowledge}, i.e., the class similarity structure encoded in the teacher logits, which encourages the student to align its features with the teacher's \citep{hinton2015distilling, allen2020towards}. Let $\phi$ be a non linear feature extractor and $h$ be an affine layer, with $z = h \circ \phi$ being the network's logits. We call $\phi(x)$ the features associated with the input $x$.

We test whether distillation leads to higher feature similarity between the distilled student and the teacher at various dataset sizes. 
Comparing teacher and student features on an individual neuron level yields inconclusive findings (\cref{feature-alignment-section}). Therefore 
we study instead the inner product across the width dimension (which is invariant to permutations of neurons), $k_{\phi}$, named \textit{feature kernel} \citep{kornblith2019similarity}:
\[k_\phi(x,x') := \langle\phi(x), \phi(x')\rangle.\]
 We can measure the similarity of two feature kernels using the Centered Kernel Alignment (CKA) \citep{kornblith2019similarity}.
We provide a brief overview of CKA in \cref{feature-alignment-section} and we refer the reader to \citep{kornblith2019similarity, cortes2012algorithms} for more details on the CKA.  


\begin{wrapfigure}{r}{0.3\textwidth}  
    \vspace{-0.2in}
    \includegraphics[width=0.3\textwidth]{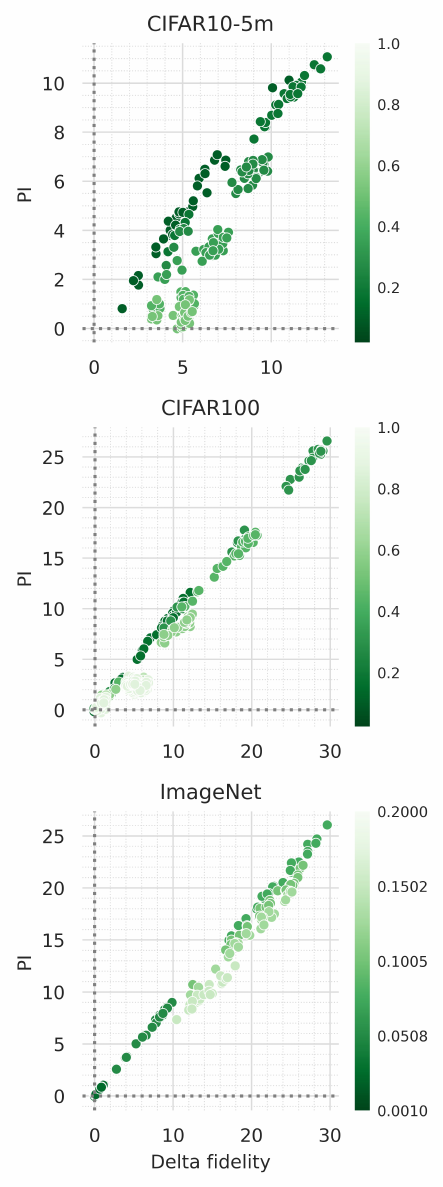}  
    \caption{\textbf{Fidelity and PI correlate.} Delta fidelity is the difference to the fidelity of an LT student trained on the same amount of data. Lighter colours correspond to higher $\kappa$. }
    \label{fidelity-test} 
    \vspace{-0.4in}
\end{wrapfigure}
\cref{feature-kernel-alignment} shows that KD induces higher feature kernel alignment than LT, particularly in low-data regimes. We observe that both the PI and the feature alignment increase as $\kappa$ decreases, suggesting a strong correlation between improved student generalization and alignment with the teacher's features.
In \cref{FKaccuracy} we {find a strong} correlation between the two across dataset sizes.

Additionally, in \cref{cross-cka}
we plot the kernel alignment between students trained with different seeds on the same input data, and
 we observe a significantly higher similarity among the KD students
compared to any other pair of trained networks.
To the best of our knowledge \emph{this is the first time logit-based distillation has been observed to result in representational alignment} 
. 
The mechanisms giving rise to this phenomenon are not trivial, given that the student only has access to the teacher logits, not features. In \cref{appssec:KDKA} we begin to investigate in this direction. 
It is worth noting that our results differ substantially between image and language data (in \cref{feature-kernel-alignment}). In the latter case, the feature kernel alignment between distilled student and teacher is often \emph{lower than the baseline}. These result suggests that there may be different mechanisms behind distillation in language settings compared to image classification. 
Although further research is needed to establish whether the different results on language and vision may be reflective of these tasks' different properties, overall these results indicate that feature learning holds promise for theoretical understanding of distillation. 



\subsection{Student (in)fidelity}
\label{fidelity-section}

Finally, we examine another widely held view on distillation: that with enough data and training, the student should eventually reproduce the teacher (perfect fidelity) \citep{beyer2022knowledge}. 
\citet{stanton2021does} observe that 
perfect fidelity is often neither attainable nor necessary to achieve good performance in practice. However, we are interested in assessing the role of fidelity at lower dataset sizes. In particular, is there a relation between the observed PIs and the degree of fidelity when $\kappa < 1$?

Following \citet{stanton2021does}, we measure fidelity using \emph{average Top-1 Agreement} 
\[\E_D[\mathbb{1}\{ \operatorname{argmax}_c (p_t(x))_c = \operatorname{argmax}_c (p_s(x))_c\}]\] 
and focus on self-distillation. Note that fidelity is distinct from feature alignment since it is measured on the outputs of the model, however high feature alignment may be a cause of high fidelity.  In contrast to \citet{stanton2021does}, we 
find a strong positive correlation between test fidelity and PI over multiple values of $\kappa$ and across datasets (\cref{fidelity-test}), despite fidelity always falling short of the $100\%$ target. {This suggests that alignment with teacher predictions may be 
a driving factor in the PI on small datasets. Thus, we may revise the conclusions of \citet{stanton2021does} stating that the bulk of the performance increment observed with distillation correlates with the alignment to the teacher, however perfect alignment is not necessary nor achieved in practical settings.}




\section{Analysis of Contributing Factors}
\label{sec:part3}
In this section, we analyze the mechanisms underlying data efficiency in distillation. We explore how components of the objective influence performance, identify the ranges of $\kappa$ that yield the largest gains, and examine when label training surpasses distillation. We also study how varying model sizes affects these patterns.

\subsection{Interplay of Model Size and Dataset Size}
\label{high-data}

\begin{figure}[h]
\centering  
{\includegraphics[width=0.8\textwidth]{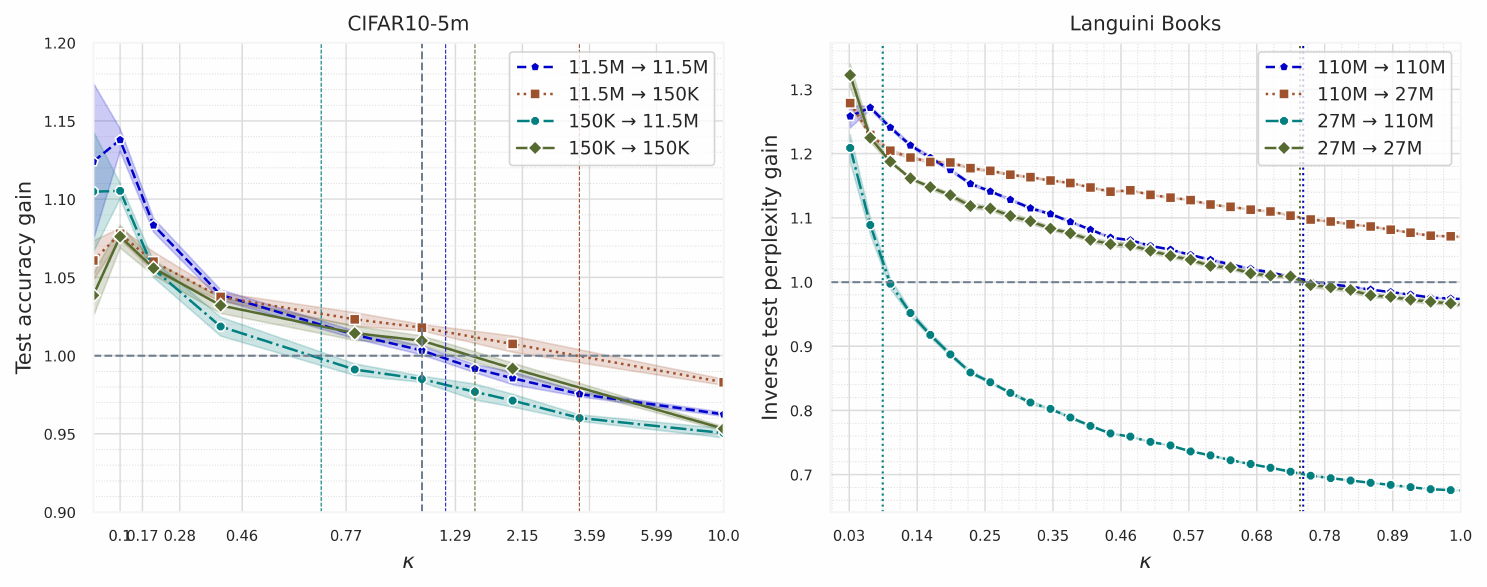}}%
\caption{\textbf{Relative size matters.} Depicted is the relative performance gain (as defined in \cref{sec:part1}) on CIFAR10-5m and Languini Books. The results are averaged over $5$ seeds. The vertical dashed lines mark the intersection point $\kappa^*$ for each configuration.}%
\label{model-scale}%
\vskip-0.1in
\end{figure}

Distillation was originally proposed to compress models without sacrificing performance, so it is common to consider teacher and student networks of different sizes. Here, we systematically vary teacher-student size combinations and dataset fractions to assess their impact on data efficiency. Specifically, we consider three cases: teacher larger than student, teacher smaller than student, and teacher equal to student. Experiments are conducted on CIFAR10 and Languini Books, with details summarized in the legend of \cref{model-scale} and configurations provided in \cref{configurations-details}. For CIFAR10-5m, we extend $\kappa$ beyond $1$ to explore behavior in the high-data regime.
From \cref{data-efficiency,model-scale}, several patterns emerge:  

\begin{enumerate}
    \item \textbf{Diminishing returns with increasing $\kappa$.}  
   The gain from distillation is most pronounced at small dataset fractions and decreases as $\kappa$ grows. Once the student is exposed to more data than the teacher (or is more expressive), label training overtakes distillation. This is consistent across both CIFAR10 and Languini Books.  

    \item \textbf{Saturation relative to teacher performance.}  
   Distilled students achieve performance slightly above the teacher for $\kappa>1$, echoing prior self-distillation findings \citep{furlanello2018born, allen2020towards, stanton2021does}. The convergence of student error close to the teacher, regardless of model size, reflects bias-variance considerations: in the high-data regime, the irreducible bias limits additional gains from distillation \citep{menon2021statistical} (we discuss this point in more detail in \cref{app-section: infidelity}).  

    \item \textbf{Effect of student-to-teacher size ratio.}  
   The value of $\kappa^\star$—where distillation and label training match in performance—is inversely correlated with $P_{Student}/P_{Teacher}$. For CIFAR10, $P_{Student}/P_{Teacher}\approx76.66$ and $\kappa^\star\approx0.7$; for Languini Books, $P_{Student}/P_{Teacher}\approx4.07$ and $\kappa^\star\approx0.083$.  

    \item \textbf{Implications for data efficiency across datasets.}  
   These observations suggest a simple relationship $\kappa^\star \propto (P_{Student}/P_{Teacher})^{-1}$, highlighting that $\kappa^\star$ essentially measures how much labeled data is needed for conventional training to reach the teacher’s performance. In self-distillation, where teacher and student are identical, $\kappa^\star$ is close to $1$. Increasing student overparameterization reduces $\kappa^\star$, implying that larger students extract more benefit from teacher guidance in low-data regimes. Importantly, despite differences in absolute values, this pattern holds consistently across both CIFAR10 and Languini Books, suggesting that the interplay between student size and dataset fraction is a robust, general phenomenon.  
   
\end{enumerate}

In summary, these findings reveal that the data efficiency of distillation is strongly modulated by the relative model size and the dataset fraction: smaller students benefit more from distillation at higher $\kappa$, while larger students require less data to match the teacher. This sets the stage for understanding how other factors, such as objective parameters, further shape distillation performance.

\subsection{Temperature and Label Smoothness}  
\label{sec:objective-parameters}

We next examine how the components of the distillation objective influence data efficiency, focusing on the role of output smoothness. In particular, we study the effect of temperature $\tau$ and the difference between soft and hard teacher targets, connecting the two experiments through the concept of label smoothness.

\begin{figure}[!ht]
\begin{center}
\includegraphics[width=0.64\columnwidth]{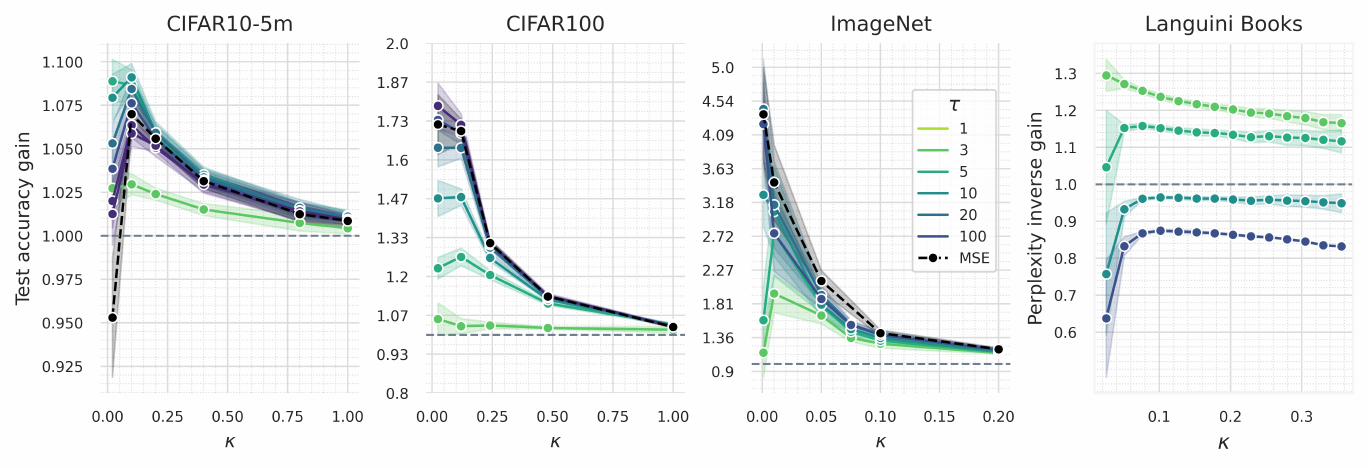}
\hfill
\includegraphics[width=0.34\columnwidth]{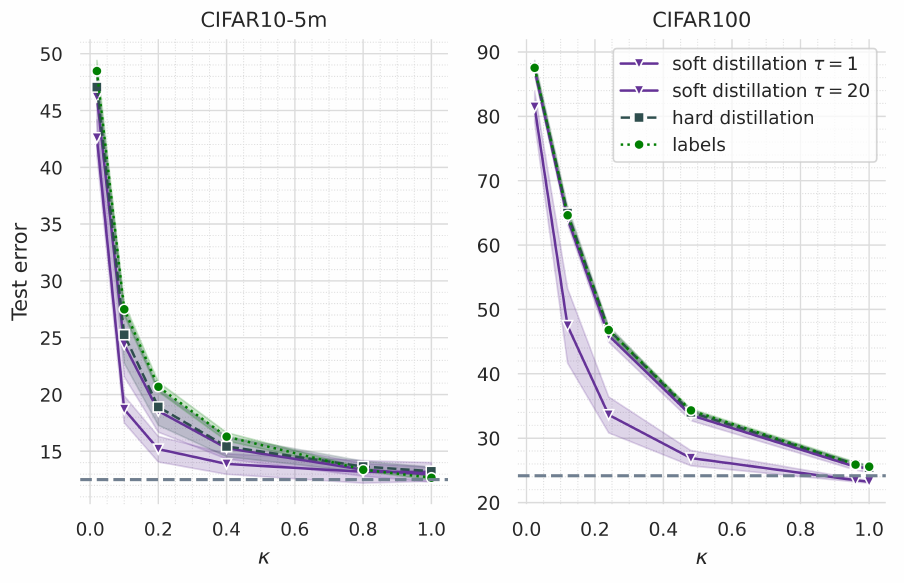}
\caption{\textbf{Impact of label smoothness on data efficiency.} (Left) Test accuracy gain of distilled students as a function of temperature. Higher temperatures produce smoother teacher distributions, increasing data efficiency, while very low temperatures reduce gains. (Right) Comparison of soft vs hard labels on CIFAR10-5m and CIFAR100, showing that retaining soft probabilities for non-target classes is necessary to achieve data-efficient distillation. Together, these experiments highlight the role of label smoothness—either via temperature or explicit soft labels—in enabling performance gains in low-data regimes.
}
\label{temperature-soft}
\end{center}
\vskip -0.2in
\end{figure}

\textbf{Temperature.}  
The temperature $\tau$ scales both the teacher and student logits before applying the softmax. Higher temperatures produce smoother label distributions, spreading probability mass over non-target classes, while lower temperatures generate peaked outputs concentrated on the top class. In the KL loss, increasing $\tau$ effectively scales the gradient by $1/\tau$, and in the limit $\tau \to \infty$ the loss approaches a squared error on the softened distributions \citep{hinton2015distilling}. As shown in \cref{temperature-soft} (left), smoother labels obtained via higher temperatures significantly improve data efficiency, particularly in low-data regimes, highlighting that the probabilistic structure of the teacher outputs is critical for effective learning.

\textbf{Soft vs. hard labels.}  
Motivated by the temperature results, we test whether removing the smoothness entirely—by replacing the teacher’s soft outputs with hard labels—impacts data efficiency. \cref{temperature-soft} (right) shows that using hard labels consistently reduces performance gains, especially for small dataset fractions. This confirms that the non-zero probabilities on non-target classes, which are emphasized at higher temperatures, are essential for transferring knowledge efficiently. In other words, the gains observed from tuning the temperature are largely due to the increased smoothness of the teacher signal; completely hard targets eliminate this benefit.

\textbf{Summary of findings.}  
Together, these experiments establish a clear link between label smoothness and data efficiency: higher temperatures create softer, more informative targets that distribute knowledge across classes, leading to stronger performance in low-data settings. Conversely, hard labels remove this information, diminishing the benefit of distillation. These results indicate that data efficiency is not merely a function of the student’s architecture or training procedure, but critically depends on the probabilistic structure of the teacher outputs and the induced optimization dynamics.

\section{Final discussion \& Conclusions}
\label{sec:discussion}

In this work, we have investigated knowledge distillation through a novel experimental framework that systematically varies the dataset size, with particular focus on the low-data regime. This approach has revealed several fundamental aspects of distillation that have received limited attention in prior literature.

Our primary finding, illustrated in \cref{data-efficiency}, is that the performance gains associated with distillation are markedly amplified when the student is trained on a reduced fraction of the dataset. In other words, distillation exhibits pronounced data efficiency in low-data regimes. Importantly, these effects are primarily of theoretical interest, as the additional computational cost introduced by teacher inference often outweighs the practical efficiency gains, highlighting a dissociation between data efficiency and computational efficiency.

By extending the analysis beyond the conventional $\kappa=1$ setting, we provide a more comprehensive characterization of the phenomenon. Observations at $\kappa=1$, which have dominated the existing literature, can now be interpreted as a special case within a broader spectrum of dataset sizes. This re-framing allows us to reconcile previously reported empirical findings with our results across varying data regimes.

We also critically evaluate several prevailing hypotheses in the distillation literature. Experiments addressing the label smoothing hypothesis indicate that the performance benefits of distillation cannot be fully explained by label regularization alone. Similarly, investigations into feature alignment and fidelity reveal that distillation induces non-trivial representational alignment and improves agreement with teacher predictions, particularly in low-data regimes, consistent with the so-called "dark knowledge" hypothesis. Nevertheless, these mechanisms appear task-dependent, with distinct behaviours observed in image classification and language modelling.

Finally, our empirical findings suggest several directions for future research. The pronounced dependence of data efficiency on factors such as temperature, label smoothness, and teacher-student size ratios points to underlying optimization dynamics that are not yet fully understood. Characterizing these dynamics theoretically could provide deeper insights into why and when distillation improves generalization, particularly in data-constrained scenarios. Overall, our results provide a unified and systematic perspective on the factors governing distillation, offering both clarification of existing observations and inspiration for further theoretical and empirical investigations.

\section*{Impact Statement}
By highlighting data efficiency as a fundamental facet of KD, our study shifts the understanding of how distillation works and opens new pathways for research. This has significant implications for improving model performance in data-scarce environments, which is crucial for fields like medical imaging, autonomous driving, and natural language processing. Our work fosters advancements in deep learning methodologies, promoting more efficient and effective deployment of AI technologies.

\bibliography{ref}
\bibliographystyle{icml2024}


\newpage
\onecolumn

\appendix
\clearpage
\addcontentsline{toc}{section}{Appendix} 
\part{Appendix} 
\parttoc 

\clearpage

\section{Experimental Details}
\label{details}

\subsection{Dataset, Networks \& Configurations}
\label{configurations-details}

We repeat our experiments on $4$ different datasets, namely CIFAR10-5m (C10) \citep{nakkiran2020cifar5m}, CIFAR100 (C100) \citep{krizhevsky2009cifar}, IMAGENET (IMN) \citep{deng2009imagenet} and \emph{Languini Books (LBOOKS)} \citep{languinikitchen} , and several networks. In particular, for the image datasets we use a set of convolutional networks and for the {LBOOKS} dataset we use GPT networks of varying sizes. An overview of the experiments configuration is given in \cref{experiments-table}. We use a publicly available extended version of CIFAR10 figuring around $6$ million images, synthetically generated by sampling from a generative model trained on CIFAR10 (commonly named CIFAR 5m). We evaluate our models on the test set also included in the CIFAR 5m collection. The dataset has been released together with the paper \citep{nakkiran2020cifar5m}. 


\begin{table}[ht]
\caption{\textbf{Overview of the experiments configurations.} The lines marked by the $^\star$ symbol refer to experiments presented in the Appendix.}
\label{experiments-table}
\vskip 0.15in
\begin{center}
\begin{small}
\begin{sc}
\begin{tabular}{lcccl}
\toprule
Dataset  & Student Networks ($P$) & Teacher Networks & SELF & Name\\
\midrule

\multirow{6}{*}{CIFAR10 (+5m)} &  \multirow{2}{*}{Vanilla CNN (150K)}& Vanilla CNN (150K)& $\surd$ & SMALL$\to$SMALL \\
 & & ResNet18 (11.5M) & $\times$ & BIG$\to$SMALL \\
 & \multirow{2}{*}{ResNet18 (11.5M)} & Vanilla CNN (150K)& $\times$ & SMALL$\to$BIG \\
 & & ResNet18 (11.5M) & $\surd$ & BIG$\to$BIG \\
 & \multirow{2}{*}{ViT (6.3M)$^\star$} & ViT (6.3M)& $\surd$ & - \\
 & & ResNet18 (11.5M) & $\times$ & - \\
 \midrule
{CIFAR100} & ResNet18 (11.5M) & ResNet18 (11.5M) & $\surd$ & - \\
 \midrule
 IMAGENET & ResNet50 (	25.6M) & ResNet50 (25.6M) & $\surd$ & - \\
 \midrule
 \multirow{6}{*}{LANGUINI BOOKS}& {GPT MINI (27M)} & GPT MINI (27M)  & $\surd$ & MINI$\to$MINI \\
 & GPT MINI (27M) & {GPT SMALL (110M)} & $\times$ & SMALL$\to$MINI\\
 & GPT MINI (27M) & {GPT MEDIUM (336M)} & $\times$ & MEDIUM$\to$MINI \\
 & {GPT MINI2 (67M)} &  GPT MEDIUM (336M) & $\times$ & MEDIUM$\to$MINI2 \\
 & GPT SMALL (110M) & {GPT MINI (27M)} & $\times$ & MINI$\to$SMALL\\
 & GPT SMALL (110M) & {GPT SMALL (110M)} & $\surd$ & SMALL$\to$SMALL\\
\bottomrule
\end{tabular}
\end{sc}
\end{small}
\end{center}
\vskip -0.1in
\end{table}

\paragraph{Exact configuration in each plot}
For the CIFAR10 and Languini Books dataset we report the network configuration used in each plot shown in the main paper: 
\begin{itemize}
    \item \cref{data-efficiency} C10: BIG$\to$BIG,  LBOOKS: SMALL$\to$SMALL.
    \item \cref{feature-kernel-alignment} C10: SMALL$\to$SMALL,  LBOOKS: MEDIUM$\to$MINI2.
    \item \cref{cross-cka} C10: SMALL$\to$SMALL
    \item \cref{model-scale} C10: all except those including ViT,  LBOOKS: MINI$\to$MINI, SMALL$\to$MINI, MINI$\to$SMALL, SMALL$\to$SMALL.
    \item \cref{temperature-soft} (Left) C10: SMALL$\to$SMALL,  LBOOKS: MEDIUM$\to$MINI2. (Right) C10: BIG$\to$BIG 
    \item \cref{fidelity-test} C10: BIG$\to$BIG 
\end{itemize}

\subsubsection{Range of $\kappa$.}
\label{rangeofkappa}

Exact set of values of $\kappa$ used for each dataset: 
\begin{itemize}
    \item C10: $[0.02      ,  0.1       ,  0.2       ,  0.4       ,  0.8       ,
        1.        ,  1.5, 2., 3.3      , 10.   
       20.]$
    \item C100: $[0.024, 0.12 , 0.24 , 0.48 , 0.96]$
    \item IMN: $[0.001, 0.01 , 0.05 , 0.075, 0.1  , 0.2  , 0.3 ]$
    \item LBOOKS: We train GPT-like language models on the Languini Books dataset in a streaming fashion, i.e. each batch is processed only once. Therefore, $\kappa$ dynamically increases during training. 
\end{itemize}

CIFAR10-5m is a synthetic dataset of similar distribution as CIFAR10 with $\sim6$M instead of $60$K samples. This allows us to investigate $\kappa \gg 1$ for teachers pre-trained on CIFAR10, as discussed in \cref{sec:part1}. In particular, we perform experiments using up to $20\times$ more data than the teacher training data with CIFAR10-5m. 

\subsubsection{Network architectures}

In line with common practice, all our networks are of the form, $f(x) = (h \circ \phi)(x)$, for non-linear feature extractor $\phi$ and linear $h$. Hereafter we may refer to $\phi$ as the network \emph{backbone} and to $h$ as the network \emph{head}. Unless stated otherwise, all the head layers take the form of a linear map from the feature space $\phi$ to the logit space $z$: $h(\zeta) = W\zeta + b$, $W$ being the weight matrix and $b$ the bias.

\paragraph{Vanilla CNN} The convolutional backbone is composed of four convolutional blocks, each consisting of a $3\times3$ convolution (with stride $1$ and padding $1$), followed by an optional \emph{batch normalisation} layer, a \emph{ReLU} nonlinearity, and a \emph{max-pooling} operation.
The number of filters doubles at each block: the first convolution uses $20$ channels, followed by $40$, $60$, and $160$ filters, respectively. The first block has no pooling, while the following three are each followed by a $2\times2$ max-pooling layer (stride $2$), and a final $4\times4$ pooling operation reduces the spatial resolution before flattening.
The resulting feature vector, of dimension $160$, is passed to a fully connected layer producing the class logits. 

\paragraph{ResNets} We reproduce the original structure of residual convolutional networks described by \citet{he2016deep}. We use a \emph{ResNet18} (feature layer width $512$) for CIFAR10 and CIFAR100, and a \emph{ResNet50} (feature layer width $1024$) for ImageNet. 

\paragraph{GPT} We use the GPT2-inspired transformer model provided in the Languini benchmark \citep{languinikitchen}. In our experiments we employ $4$ GPT2 models of different sizes. In particular, the width and depth (measured in number of \emph{attention blocks}) of the backbone changes between sizes, but all the models share the same block type. The code of the Languini library is publicly available on GitHub\footnote{ \href{https://github.com/languini-kitchen/languini-kitchen/tree/main}{https://github.com/languini-kitchen}}. The \emph{MINI} GPT network has width $512$ and depth $4$; the \emph{MINI2} GPT network has width $1024$ and depth $4$; the \emph{SMALL} GPT network has width $768$ and depth $6$; and finally the \emph{MEDIUM} GPT network has width $1024$ and depth $24$. We use two trained MINI and MEDIUM networks as teachers.

\subsection{Training procedures}

All our experiments involve two training steps. First, we train one teacher network on the full dataset (or a fixed portion thereof in case of C10 and LBOOKS data). Second, we train another network (the student) on a variable portion of the dataset. 

\paragraph{Teachers}

We train one teacher for C100 and IMN, two teachers for C10 and three teachers for LBOOKS. The seed of the teacher is fixed and once trained we use the teacher as a black-box function. Importantly, the teachers are trained with one-hot-labels following common practices (see \cref{hyperparameters} for details). 

The C100 and IMN teachers are trained on the full training set. The C10 teachers are trained on a fixed random sample of $60$K images from the almost $6$M available samples. To ease comparison, the LBOOKS teachers are trained on the same amount ($\approx 8.3 G$) of tokens. 

\paragraph{Students}

For each experimental configuration, we train two identical networks (which we call students) \emph{with identical training settings}, either using one-hot-labels or soft-label targets provided by the teacher. Each experiment is repeated over $5$ seeds, which means a total of $10$ networks (with $5$ different initialisations). For each dataset, we train these $10$ networks on multiple fractions of data (identified by the value $\kappa$, see \cref{rangeofkappa} above). Moreover, we distil all students with different temperatures $\tau$ (see \cref{hyperparameters} for the list). 

Notice that a student trained with one-hot labels on the full dataset ($\kappa=1$) is equivalent to the teacher (up to its initialisation). For this reason, we keep the same training setup for teachers and students. Moreover, we do not change training hyperparameters between label training and teacher distillation to allow for a better comparison. 

\subsubsection{Hyperparameters}
\label{hyperparameters}

We repeat all of our experiments over $5$ seeds, which affect the network initialisation and the data sampling processes. Moreover, we vary the temperature of distillation in the range $[0.1, 1, 3, 5, 10, 20, 100]$, and we simulate the case $\tau \to \infty$ with an $l_2$ loss on the logits (cf \citep{hinton2015distilling}). Finally, unless stated otherwise, we use the SGD optimiser for training. 

For C10 we do not use optimal training hyperparameters. Therefore, the performance achieved by teacher and student networks is not maximal with respect to their capacity. For all the other datasets, however, we rely on publicly available optimal "training recipes" which have been tuned to the architecture. Therefore in the case of C100, IMN and LBOOKS the performance of our models is high relative to the model capacity.

\paragraph{CIFAR10} 
For both the teacher and the student networks pair we use the following training hyperparameters: learning rate $= 0.1$, with a linear warmup over the first $5$ epochs and subsequently annealing the learning rate with a cosine schedule, weight decay $=0.001$, batch size $256$, $30$ epochs. We use random augmentations consisting of crops to $32\times 32$ and horizontal flips.

\paragraph{CIFAR100} For both the teacher and the student networks pair we use the following training hyperparameters: learning rate $= 0.1$, with a linear warmup over the first epoch and subsequently reducing the learning rate by a factor of $5$ after $60,120$ and $160$ epochs, weight decay $=0.0005$, momentum $= 0.9$, batch size $128$, $200$ epochs. We use random augmentations consisting of crops to $32\times 32$, horizontal flips and rotations of $15$ degrees maximum.

\paragraph{IMAGENET} For both the teacher and the student networks pair we use the following training hyperparameters: learning rate $= 0.1$,  reducing the learning rate by a factor of $10$ every $30$ epochs, weight decay $=0.001$, momentum $= 0.9$, batch size $64$, $90$ epochs. We use random augmentations consisting of crops to $224\times 224$ and horizontal flips.

\paragraph{LANGUINI BOOKS} For each GPT model we follow the standard training recipe provided by the Languini library, including Adam \cite{kingma2017adam} (cf the code for details). Importantly, we decay the learning rate at every step and always use a batch size of $128$. The \emph{MINI} teacher has been trained on $3.2$B tokens and the \emph{MEDIUM} teacher has been trained on $5.7$B tokens from the same source. 

\paragraph{Label smoothing} In our label smoothing experiments on C100 we use the same hyperparameters as \citet{yuan2020revisiting} for better comparison (although they use a different student-teacher network configuration). We then repeat the experiment on C10 (this dataset is not present in \citet{yuan2020revisiting})  using the same hyperparameters. Specifically, we set $a=0.99$ and $\alpha=0.9$ (so the distillation weight is $0.1$). Moreover, we explore $3$ temperature values, namely $\tau=1,20,100$.

\subsubsection{Compute resources} 
\label{compute-sec}
We perform all of our experiments on graphic cards NVIDIA 4090, with 24GB of GPU memory. For the larger language experiments which require higher GPU memory we parallelise our experiments over multiple devices. The maximal runtime of a single experiment is 5 days and 22 hours. The total recorded compute for the entire project (so including failed and omitted experiments)  is 1080 days. 

\newpage

\section{Additional Experiments}
\subsection{DED in Vision-Transformers.}
\begin{figure}
    \centering
    \includegraphics[width=0.7\columnwidth]{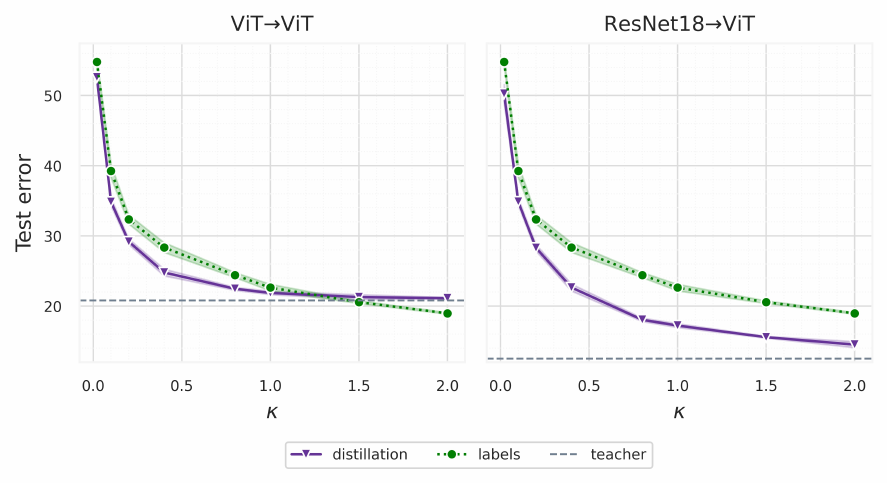}
    \caption{\textbf{DED can be observed in attention-based architectures.} Test error on CIFAR10-5m as a function of the relative training dataset size $\kappa$ for ViT models. Compared are models obtained through label training and distillation from a ViT teacher (left) and a ResNet18 teacher (right). Importantly, we observe data efficiency also for attention-based architectures when using distillation. 
    }
    \label{fig:vit}
\end{figure}

Out of curiosity and completeness in our empirical analysis we run an experiment using Vision Transformers (ViT) on CIFAR10-5m. Given that ViTs are notoriously data inefficient and the CIFAR10 dataset is relatively small, the ViT teacher we use (adapted from this \href{https://github.com/omihub777/ViT-CIFAR/tree/main?tab=readme-ov-file}{Pytorch implementation} of \citep{dosovitskiy2020image}, without extra data augmentations for better comparisons with CNNs) only achieves 80\% validation accuracy on CIFAR10. Therefore, we also compare the setting of training ViT students with the ResNet18 teacher. In \cref{fig:vit} we plot the test error of distillation and label training as we vary the fraction of training data $\kappa$. Interestingly, the performance increment is consistently higher when using the ResNet18 teacher, and it carries over the $\kappa=1$ threshold. We suspect that the reason for this difference lies in the markedly lower test error in the ResNet18 teacher, however, further experiments are needed to finalise this claim.

\subsection{Zooming into (in)fidelity.}
\label{app-section: infidelity}

\begin{figure}[ht]
\begin{center}
\centerline{\includegraphics[width=0.8\columnwidth]{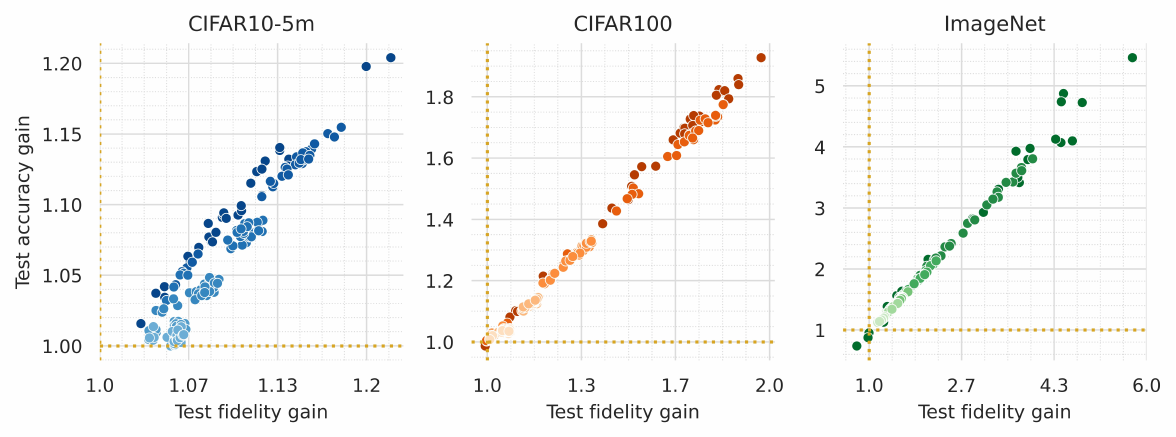}}
\caption{\textbf{Test fidelity and test accuracy correlate} Test fidelity and test accuracy over three datasets. Different points correspond to different seeds and values of $\kappa$.}
\label{fidelity-test}
\end{center}
\vskip -0.2in
\end{figure}

\begin{figure}[ht]
\vskip 0.2in
\begin{center}
\centerline{\includegraphics[width=0.8\columnwidth]{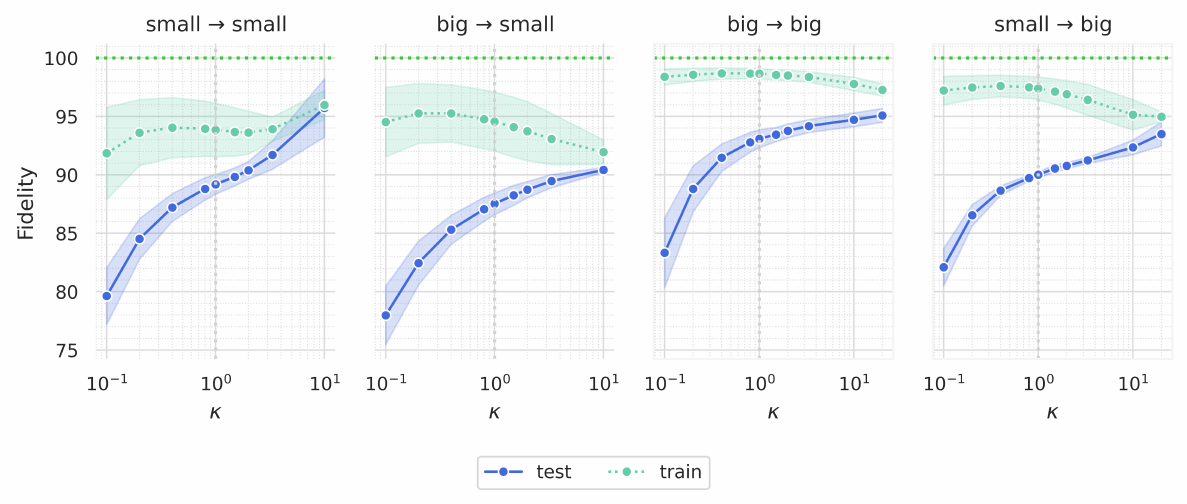}}
\caption{\textbf{} Distillation fidelity over CIFAR10-5m train and test data for different network configurations.}
\label{fidelity-cifar10}
\end{center}
\vskip -0.2in
\end{figure}

We report additional results on distillation fidelity for the CIFAR10-5m dataset, which allows us to explore the particularly interesting high-data regime. 
In \cref{fidelity-cifar10} we plot distillation fidelity on train and test data for different student-teacher network configurations. 

We must remark that several aspects of this setting are sub-optimal and do not match the experiments in \citet{stanton2021does}, therefore the conclusions must be taken with a grain of salt. To begin with, the training hyperparameters are not optimised and they are especially inadequate for the 'small' networks. Another factor which may be entangled in these results is the presence of augmentations. We adopt the same augmentations for all network configurations, despite the differences in representational capacity. Finally, in some settings, there is an irreducible approximation error due to the mismatch of student and teacher architecture, which may be a confounder to higher fidelity error. 

\begin{figure}[ht]
\vskip 0.2in
\begin{center}
\centerline{\includegraphics[width=0.8\columnwidth]{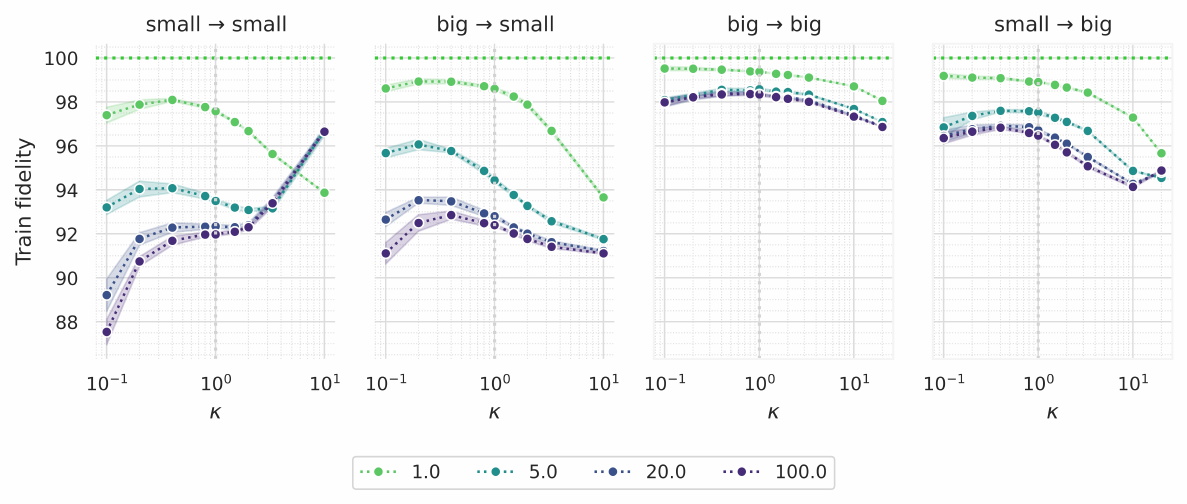}}
\caption{\textbf{Temperature affects train fidelity} Distillation fidelity  over CIFAR10-5m train data as we vary the distillation temperature $\tau$.}
\label{fidelity-temperature}
\end{center}
\vskip -0.2in
\end{figure}

Nevertheless, we observe an interesting trend in the high data regime. The train and test curves converge to the same value as $\kappa$ increases. In line with the observations of \citet{stanton2021does}, fidelity seems to converge to a value below 100\%, even when the teacher is smaller than the student. We plot the difference between train and test fidelity as a function of $\kappa$. Curiously, we find that, across all configurations, the difference curves are well approximated by $O(1/\sqrt{\kappa})$.
Further, in \cref{fidelity-temperature} we show train fidelity for multiple distillation temperatures. Temperature appears to have a strong influence on train fidelity. One hypothesis is that this effect is a consequence of the different training dynamics due to the temperature scaling the gradient. More surprisingly, the trend is reversed with respect to generalisation: higher temperatures deliver higher generalisation and lower train fidelity.

\begin{figure}[ht]
\vskip 0.2in
\begin{center}
\centerline{\includegraphics[width=0.8\columnwidth]{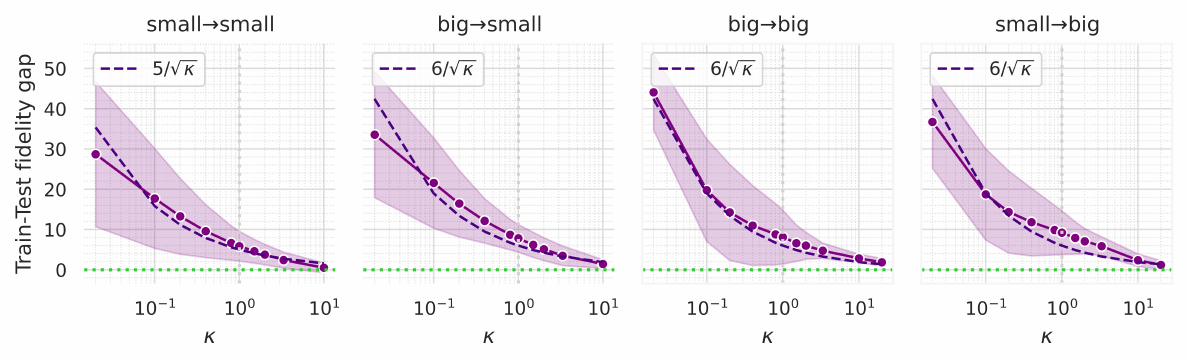}}
\caption{\textbf{The difference between train and test fidelity reduces at a $1/\sqrt{\kappa}$ rate.} We plot the difference between train and test fidelity on CIFAR10 for each network configuration. We juxtapose each curve with the best fitting $\omega/\sqrt{\kappa}$ line.}
\label{fidelity-rates}
\end{center}
\vskip -0.2in
\end{figure}

To better understand the effect of the dataset size on DED we turn to a simple bias-variance decomposition of the expected error, in a similar spirit as \citep{menon2021statistical}. 
Let $p_s(D)$ be a student trained on the dataset $D$ and $\Bar{p}_s^M$ be the mean student trained with $M$ samples, i.e. $\Bar{p}_s^M = \E_{D\sim \mathcal{P}^M}[p_s(D)]$. Taking $p_y$ to be the true label distribution\footnote{Note that by using $p_y$ instead of $\delta(y)$ we get rid of potential label noise.}, the expected squared loss $l_2(f,g) = \E_{x,y}[\|f(x) - g(x)\|^2]$ decomposes into two terms: 
\begin{equation}
\label{bias-variance}
    \begin{aligned}
        \E_{D\sim \mathcal{P}^M}[l_2(p_s(D), p_y)] =  \underbrace{\E_{D}[l_2(p_s(D), \Bar{p}_s^M)]}_\text{Variance} + \underbrace{l_2(\Bar{p}_s^M, p_y)}_{\text{Bias}^2} + \; \epsilon
    \end{aligned}
\end{equation}
where $\epsilon$ is an irreducible approximation error.
As the number of training samples grows $M\to\infty$, the variance term reduces up to the noise inherent in the optimisation process. Consequently, the bias term controls the behaviour in the high-data regime for both distillation and label training. In the case of distillation with a  fixed teacher trained on finite data, the bias term converges to a constant, which depends on the teacher accuracy on $\mathcal{P}$, as well as the bias implicit in the optimisation procedure. Thus, in the high-data regime, the positive bias penalises distillation over ground-truth targets. By the same token, when the data is scarce the variance term may be significantly higher than the bias and dominate the error. Therefore the high performance in low data regimes suggests that distillation has a variance reduction effect on the estimator, which compensates for the higher bias. And this effect is consistent across datasets and models. 

\subsection{More on feature learning.}

\subsubsection{What impact does the linear head have on feature learning?}
\label{app-sec:linear-head}

We assess the relevance of the linear \emph{head} $h$ in DED. In other words, we ask: 
\begin{quote}
    {is the observed data efficiency dependent on the linear map $h$?} 
\end{quote}
\begin{figure}
    \begin{center}
\centerline{\includegraphics[width=0.65\columnwidth]{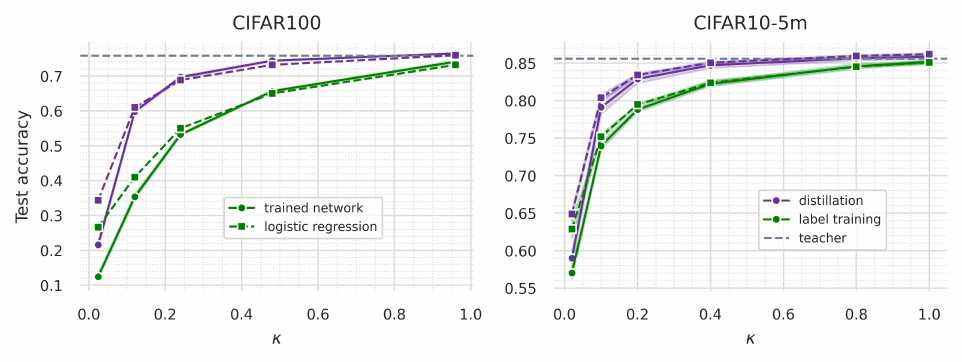}}
\caption{\textbf{Data efficiency does not depend on the linear head.} Test classification accuracy (in a 0-1 scale) as a function of $\kappa$. We compare the trained network to a logistic regression classifier (dashed lines). Label-trained students are shown side by side with distilled students (e.g. \cref{feature-alignment}).}
\label{linear-head}
\end{center}
\end{figure}

This is a natural question to ask because different feature extractors $\phi$ are known to perform differently when $h$ is trained on little data, depending on the eigendecomposition of $\phi$ \citep{bordelon2020spectrum,pmlr-v162-he22c}. 
To answer this question, we take feature extractors from teacher-distilled and label-trained students, on various fractions $\kappa$ of data, and fit a logistic regression classifier on the feature-based representation \emph{of the whole dataset} ($\kappa=1$). By fitting the \emph{linear probe} \cite{alain2016understanding} on the full dataset we are accounting for potential effects of data scarcity on the linear map $h$.

In \cref{linear-head} we show the results.  Crucially, we observe that retraining the linear layer \emph{preserves} the gain in test-accuracy of distillation and the effect of temperature (\cref{linear-head-temp}) across students, with the largest gains for small $\kappa$ as expected. We therefore conclude that the data efficiency of distillation cannot be captured wholly through the linear layer $h$ and one must consider also the network features.

\subsubsection{Does distillation induce the same features?}
\label{feature-alignment-section}
\begin{figure}[ht]
\vskip 0.2in
\begin{center}
\centerline{\includegraphics[width=0.8\columnwidth]{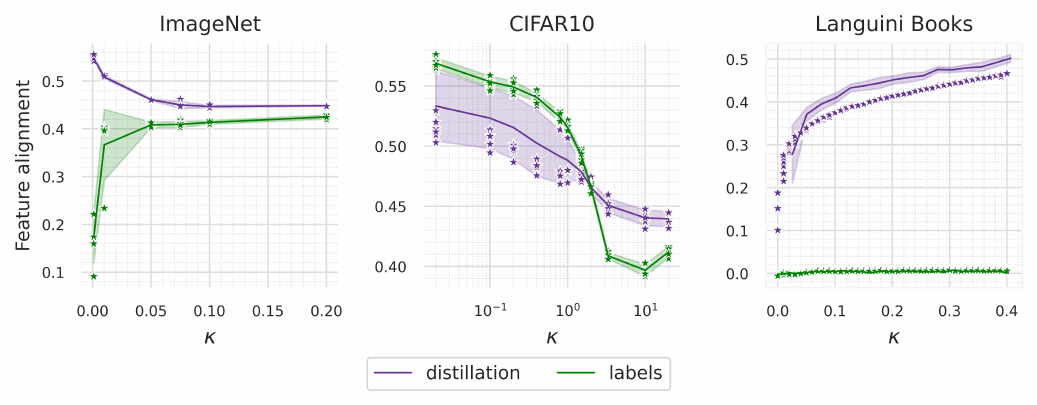}}
\caption{\textbf{Feature alignment varies across datasets and architectures.} Feature alignment (\cref{eq-feature-alignment}) between the teacher and the distillation- (purple) and label- (green) trained students as a function of $\kappa$. The markers show individual samples, and the lines represent the average. }
\label{feature-alignment}
\end{center}
\vskip -0.2in
\end{figure}


We proceed to explore the effect of distillation on the student network features $\phi$. We ask the following simple question: \emph{do the distillation-trained features approximate the teacher features?}

In order to answer this question we look at the normalised inner product between the students and teacher features when the two networks are identical. More precisely, let $a,b$ be two different instances of the same network, we define their \emph{feature alignment} to be: 
\begin{equation}
\label{eq-feature-alignment}
    \operatorname{FA}(a,b) = \frac{1}{Z} \langle \phi_a, \phi_b \rangle_D
\end{equation}
The sign $\langle \cdot, \cdot \rangle_D$ denotes the average over the data distribution, which we approximate by an average over the test set, and $Z = \sqrt{\langle \phi_a, \phi_a\rangle_D \cdot \langle \phi_b, \phi_b\rangle_D }$ normalises the score.

\begin{table}[t]
\caption{\textbf{Feature alignment does not depend on initialisation.} This table reports feature alignment averaged over several values of $\kappa$ for $5$ seeds.}
\label{FA-table}
\vskip 0.15in
\begin{center}
\begin{small}
\begin{sc}
\begin{tabular}{lccc}
\toprule
Network  & $\operatorname{FA}$ & Distillation & Same init\\
\midrule
\multirow{4}{*}{RN18} & $0.49 \pm 0.03$ & $\surd$ & $\times$  \\
 & $0.40 \pm 0.06$ & $\surd$ & $\surd$  \\
 & $0.51 \pm 0.07$ & $\times$ & $\times$  \\
 & $0.52 \pm 0.07$ & $\times$ & $\surd$  \\
 \midrule
\multirow{4}{*}{CNN} & $0.78 \pm 0.01$ & $\surd$ & $\times$  \\
 & $0.79 \pm 0.01$ & $\surd$ & $\surd$  \\
 & $0.84 \pm 0.01$ & $\times$ & $\times$  \\
 & $0.84 \pm 0.01$ & $\times$ & $\surd$  \\

\bottomrule
\end{tabular}
\end{sc}
\end{small}
\end{center}
\vskip -0.1in
\end{table}

\cref{feature-alignment} shows the feature alignment between the students and the teacher on $3$ benchmarks of different difficulty.  Importantly, feature alignment can only be computed if the teacher and student features are of the same dimension. Thus we apply this test only to the self-distillation settings. 
We do not observe a shared trend among the benchmarks, suggesting that distillation does not necessarily imply feature alignment. Note that for convolutional networks the features are taken after ReLU activation and thus the alignment will be positive. This is not the case in the transformer network. Perhaps surprisingly, we observe low alignment also when the student and teacher initialisation coincide (\cref{FA-table}).

\subsubsection{NTK alignment}
It is natural to inquire whether the alignment observed at the feature layer propagates back through the network backbone. In order to do this we look at the Neural Tangent Kernel (NTK) \citep{jacot2018neural}, a model of training dynamics in wide NNs that is exact in the infinite-width limit under certain parameterisations. In the NTK setting, an NN $f_{\theta}$ evolves as a linear model in its parameters $\theta$, with a \textit{fixed} feature map determined by its Jacobian $\frac{\partial f_{\theta}}{\partial \theta}$ at initialisation, which captures features from all layers in the NN. 

Importantly, the (last layer) feature kernel appears in the NTK computation as one summand in a sum over the network layers, because the Jacobian of $f_{\theta}$ with respect to the last linear layer is precisely the feature vector $\phi$. Therefore the NTK alignment between two networks captures offers an overview of the alignment of the feature at all the intermediate layers.

\begin{figure}[ht]
\vskip 0.2in
\begin{center}
\centerline{\includegraphics[width=0.6\columnwidth]{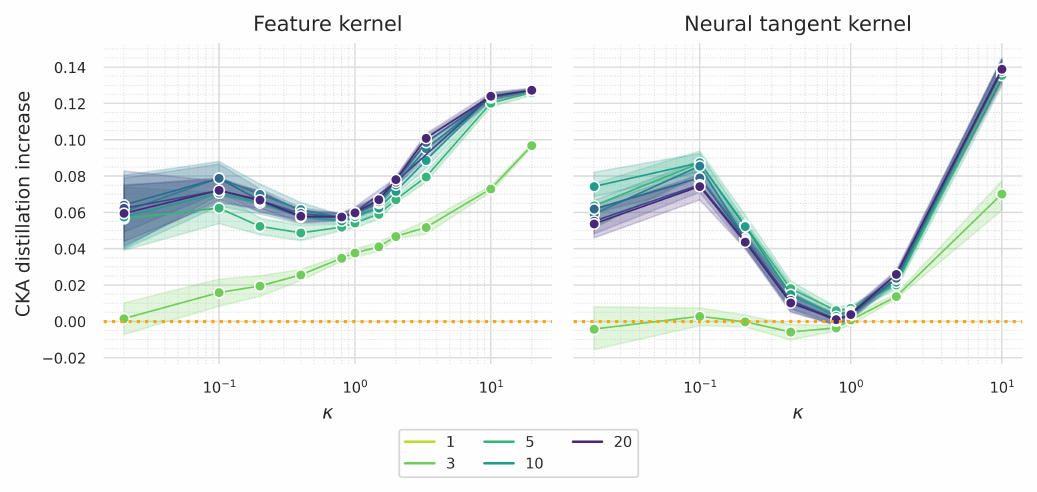}}
\caption{\textbf{NTK vs FK alignment.} The kernels are measured on CIFAR10 for the SMALL$\to$SMALL network configuration. }
\label{ntk-alignment}
\end{center}
\vskip -0.2in
\end{figure}

We compute the NTK of teacher and student networks (both distillation and labels) and evaluate their alignments using CKA. We plot the result in \cref{ntk-alignment}, alongside the feature-kernel alignments for the same experimental setting. Predictably, we observe a similar trend in the two curves. However, the feature-kernel alignment is generally higher than the NTK's, suggesting that the effect of distillation is best observed in the feature layer.

\subsubsection{Does distillation yield feature kernel alignment?}
\label{appssec:KDKA}

First, the CKA is defined as follows: 
\begin{equation}
    \label{cka}
    \text{CKA}(k_s,k_t) = \frac{\text{HSIC}(k_s,k_t)}{\sqrt{\text{HSIC}(k_s,k_s) \cdot \text{HSIC}(k_t,k_t)}}
\end{equation}
with $\text{HSIC}(k_s,k_t) = (n-1)^{-2}\cdot \text{Tr}(k_s H k_t H)$, and $H$ being a centering matrix.

We begin by looking at the case of an optimal distillation student $f_s^\star$. Say that $f_s^\star(x) = f_t(x)$ for all $x \in D_M$, ($D_M$ being the training dataset of size $M$). If we define the \emph{target kernel} as:
\begin{equation}
    \label{eq:targtker}
    k_f^{tg}(x,x') := \langle f(x), f(x')\rangle
\end{equation}
it is obvious to conclude that distillation entails equivalence of the teacher and student's target kernels on the training data (cf \citet{tang2020understanding} for evidence of this effect).  However, it is not obvious how feature kernel alignment may ensue from target kernel alignment.
Rewriting $f_s^\star(x)$ as $W_s \phi_s(x)$ and $f_t(x)$ as $W_t\phi_t(x)$ the target kernel is $k_{f}^{tg}(x,x') = \phi_s(x)^\intercal [W_s^\intercal W_s]\, \phi_s(x')$. Thus from the equivalence of target kernels, it follows that: 
\begin{equation*}
    \begin{aligned}
        \phi_s(x)^\intercal [W_s^\intercal W_s]\, \phi_s(x') = \phi_t(x)^\intercal [W_t^\intercal W_t]\, \phi_t(x')
    \end{aligned}
\end{equation*}
If $W_s$ and $W_t$ are orthogonal matrices, we can immediately conclude that the student and teacher feature kernels are equivalent up to some scaling factor. 

But in general $W_s$ and $W_t$, will not be square matrices and cannot be orthogonal. Indeed, for image classification settings we will have the output projection down to a smaller number of classes than width, and for language modelling transformers we have the opposite (the Languini vocabulary is 16k). For the image classification setting, we can hope to recover some structure in the feature and weight spaces due to the Neural Collapse phenomenon \cite{papyan2020prevalence,kim2024fixed}, which will tells us that the features and the weights in trained classification NNs on small numbers of classes will become aligned. They will also exhibit a Simplex Equiangular Tight Frame behaviour in the final layer, where class inputs are mapped to the class centroid. Investigating if Neural Collapse can help to explain the feature alignment we observe with distillation provides an interesting direction for future work.

\begin{figure}[ht]
\vskip 0.2in
\begin{center}
\centerline{\includegraphics[width=0.4\columnwidth]{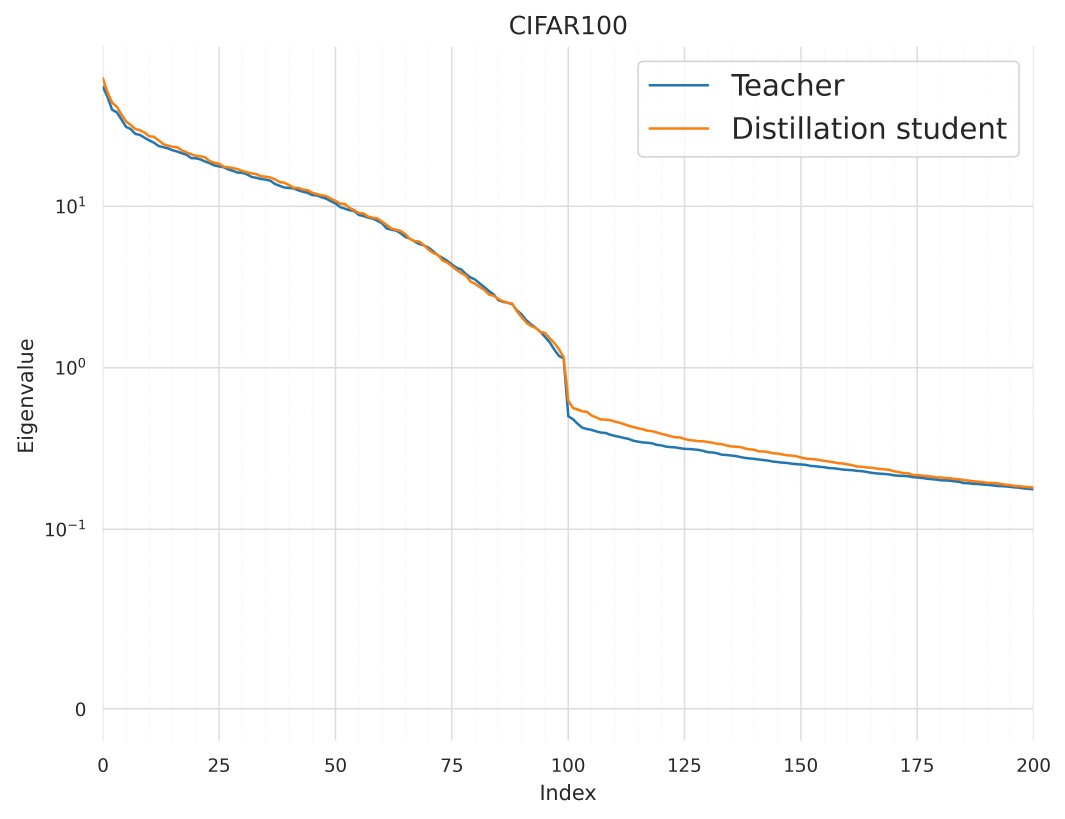}}
\caption{Eigenspectrum for teacher and a distillation student network trained on CIFAR100. We observe a drop after the first $100$ dimensions, which is often indicative of neural collapse.}
\label{C100-FK}
\end{center}
\vskip -0.2in
\end{figure}

\newpage

\section{Additional Figures and Empirical Substantiation}
\label{AdditionalPlotsFK}
This subsection includes placeholder figures for concepts discussed in the main text, for which specific existing figures were not available or suitable for direct inclusion in the main body.
\begin{figure}[ht]
\vskip 0.2in
\begin{center}
{\includegraphics[width=0.4\columnwidth]{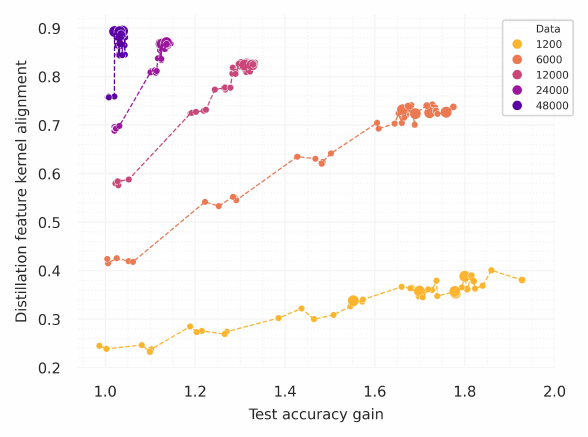}}{\includegraphics[width=0.4\columnwidth]{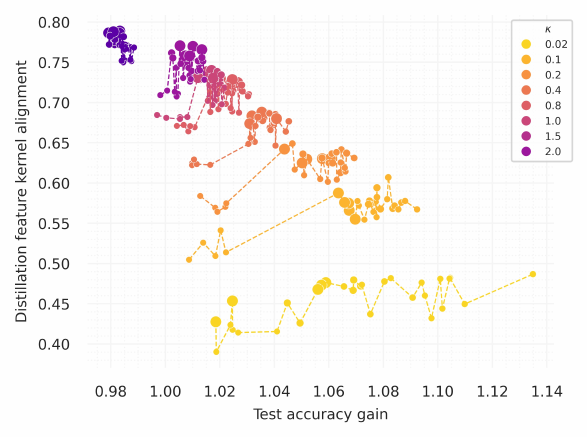}}
\caption{\textbf{Feature kernel alignment correlates with test accuracy gain.} Each point represents a different student-pair instance for varying $\kappa$ (represented by the colour) and $\tau$ (represented by the size) on CIFAR100 (left) and CIFAR10 (right). The dashed lines connect points with the same $\kappa$ to highlight the differences within equivalent data regime groups.}
\label{FKaccuracy}
\end{center}
\vskip -0.2in
\end{figure}

\begin{figure}[ht]
\vskip 0.2in
\begin{center}
\centerline{\includegraphics[width=0.5\columnwidth]{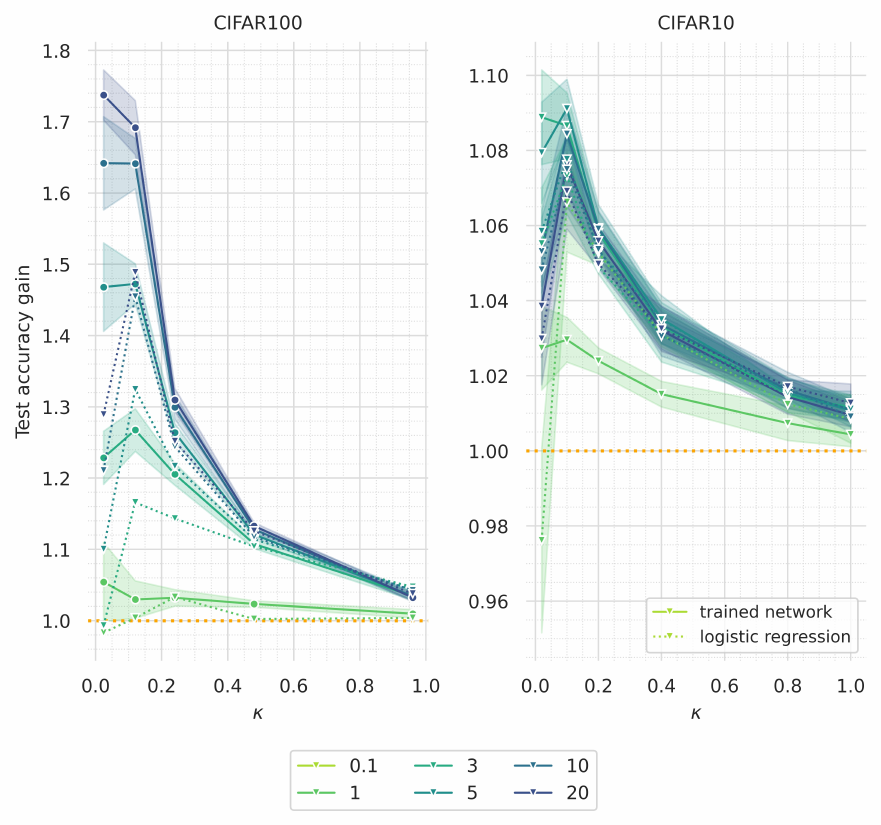}}
\caption{\textbf{Data efficiency does not depend on the linear head (2).} Test accuracy gain as a function of $\kappa$ and the distillation temperature $\tau$. We compare the trained network to a logistic regression classifier.}
\label{linear-head-temp}
\end{center}
\vskip -0.2in
\end{figure}

\begin{figure}[ht]
\vskip 0.2in
\begin{center}
{\includegraphics[width=0.3\columnwidth]{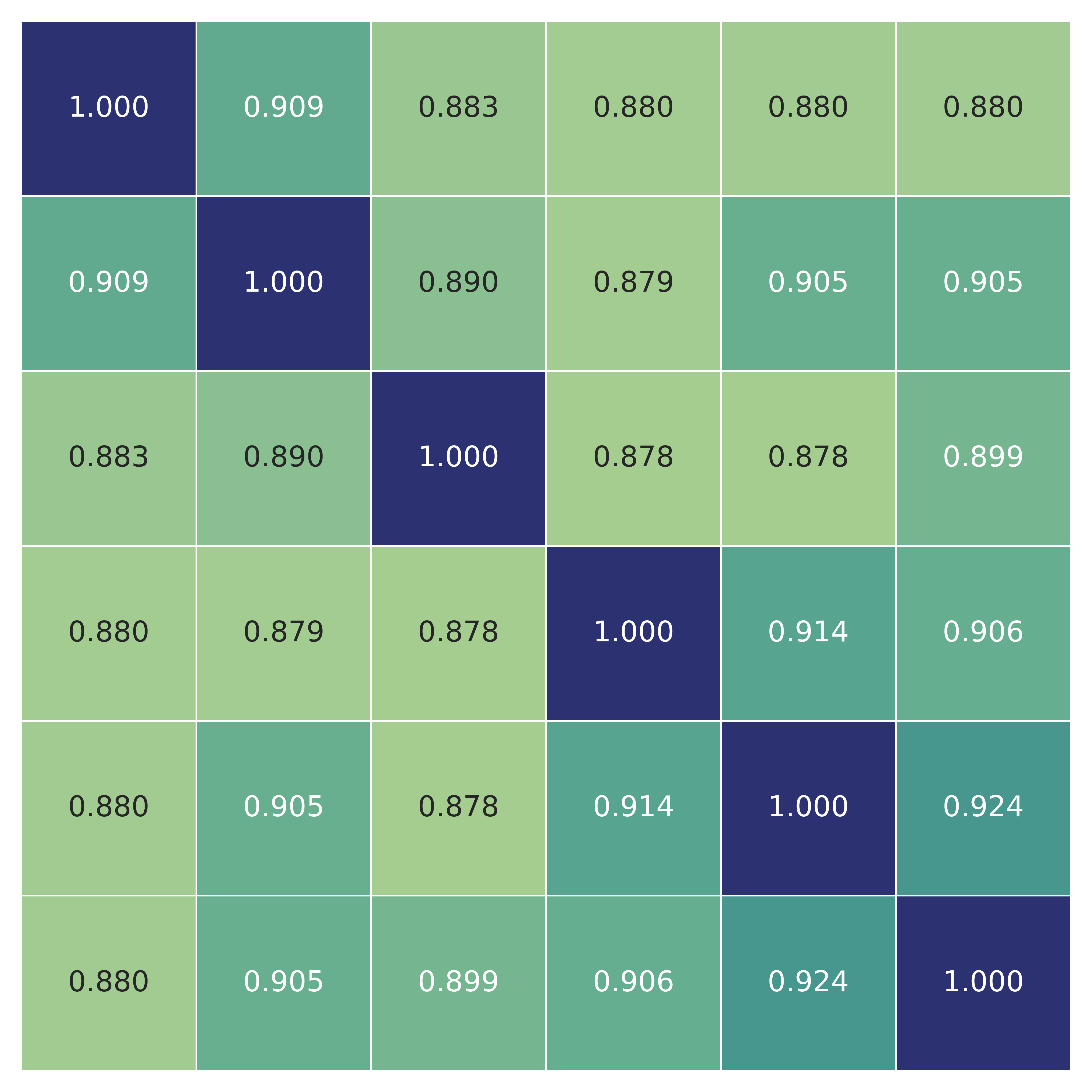}}
{\includegraphics[width=0.3\columnwidth]{figures/cross_CKA_1200_20.png}}
\caption{$\kappa=0.02, \tau=1$ (left) and $\tau=20$ (right).}
\label{cross-cka-extra}
\end{center}
\vskip -0.2in
\end{figure}

\begin{figure}[ht]
\vskip 0.2in
\begin{center}
{\includegraphics[width=0.3\columnwidth]{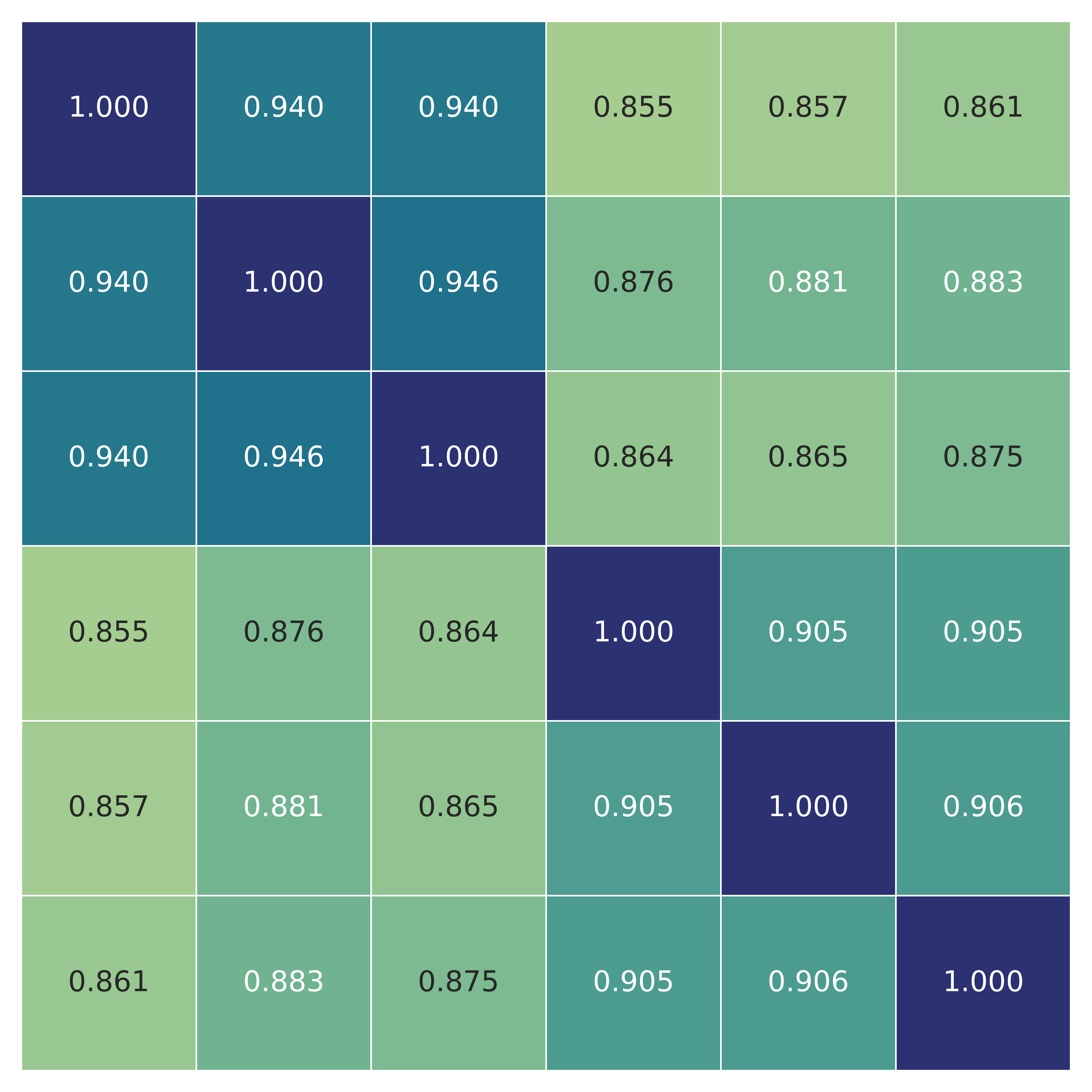}}
{\includegraphics[width=0.3\columnwidth]{figures/cross_CKA_6000_20.png}}
\caption{$\kappa=0.1, \tau=1$ (left) and $\tau=20$ (right).}
\label{cross-cka-extra}
\end{center}
\vskip -0.2in
\end{figure}

\begin{figure}[ht]
\vskip 0.2in
\begin{center}
{\includegraphics[width=0.3\columnwidth]{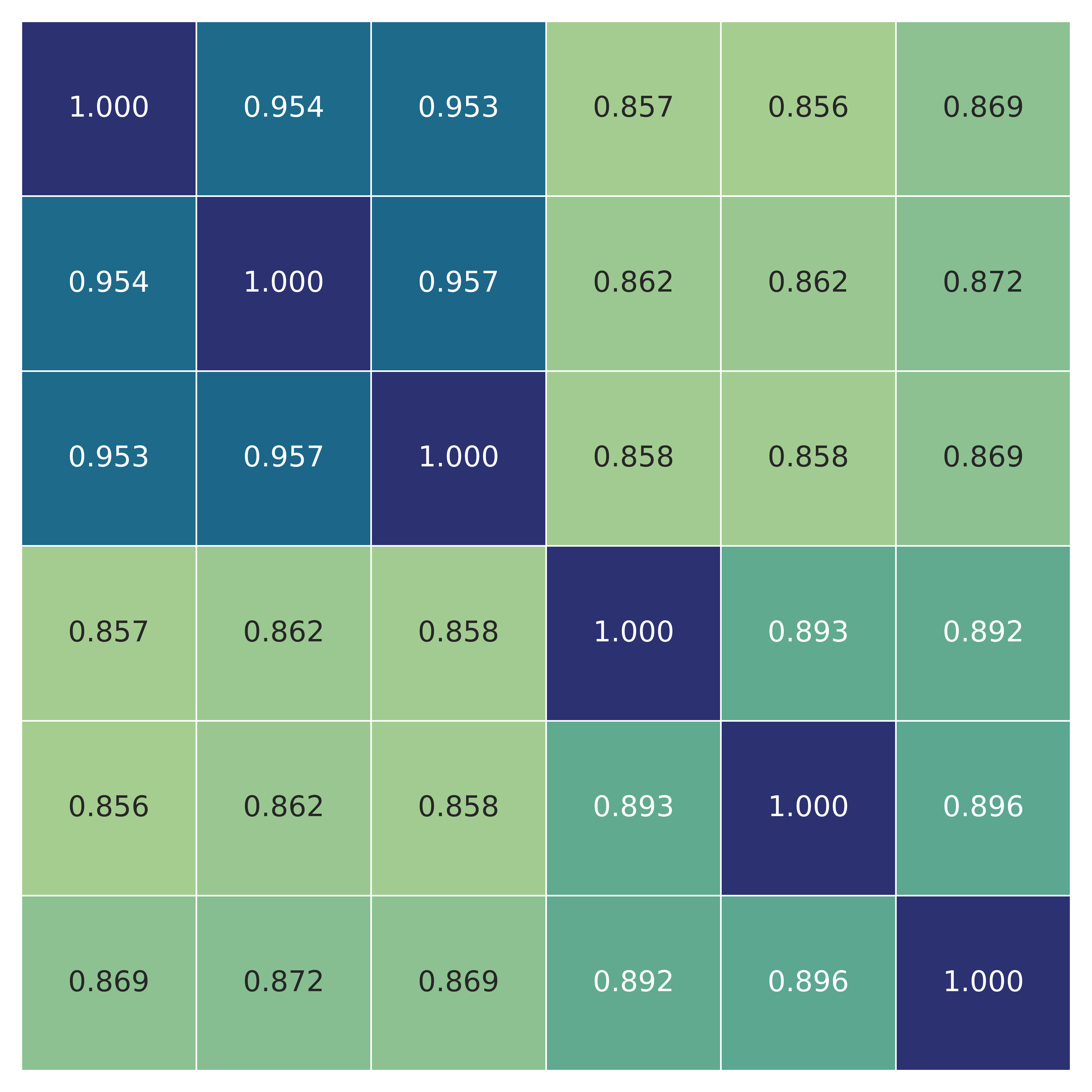}}
{\includegraphics[width=0.3\columnwidth]{figures/cross_CKA_12000_20.png}}
\caption{$\kappa=0.2, \tau=1$ (left) and $\tau=20$ (right).}
\label{cross-cka-extra}
\end{center}
\vskip -0.2in
\end{figure}

\begin{figure}[ht]
\vskip 0.2in
\begin{center}
{\includegraphics[width=0.3\columnwidth]{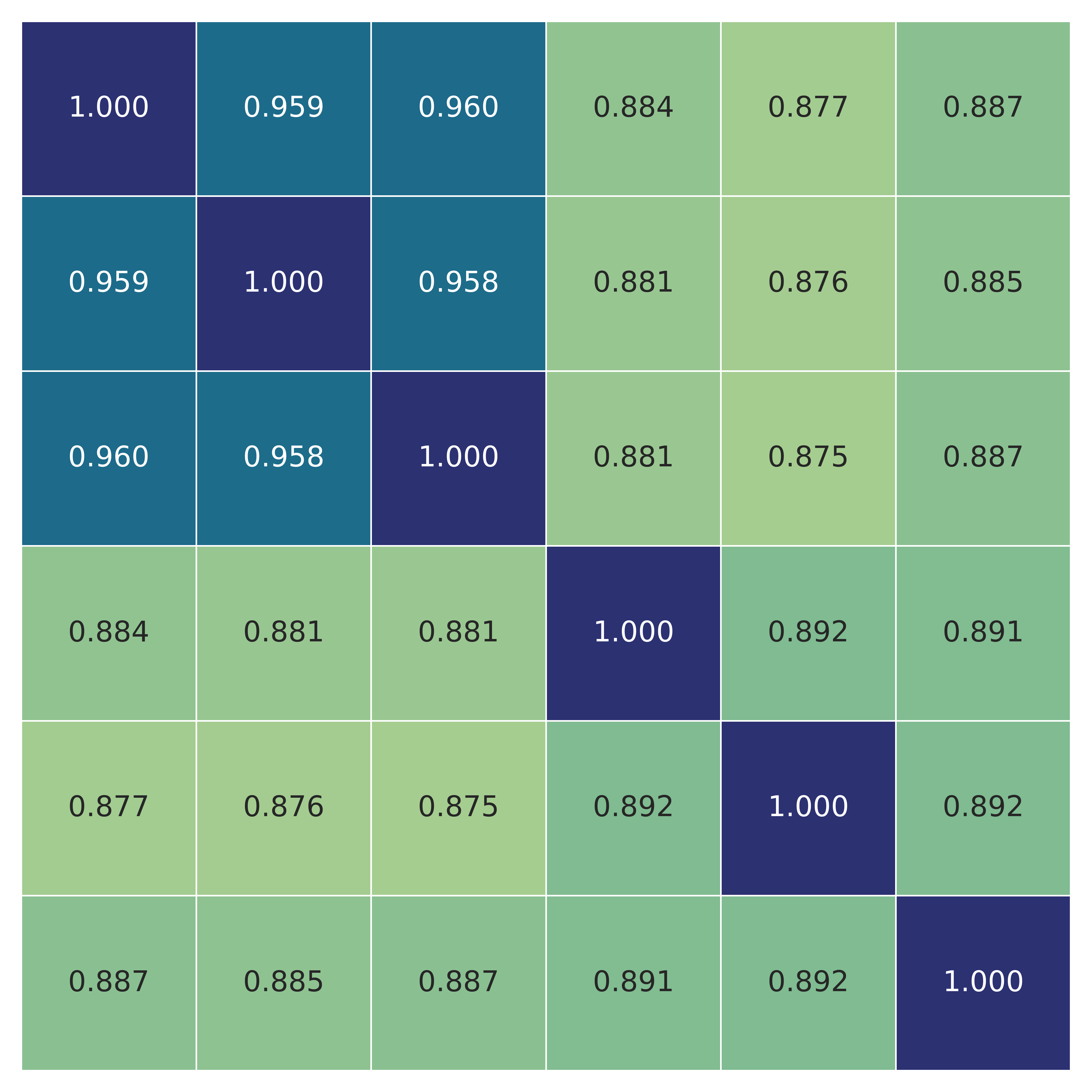}}
{\includegraphics[width=0.3\columnwidth]{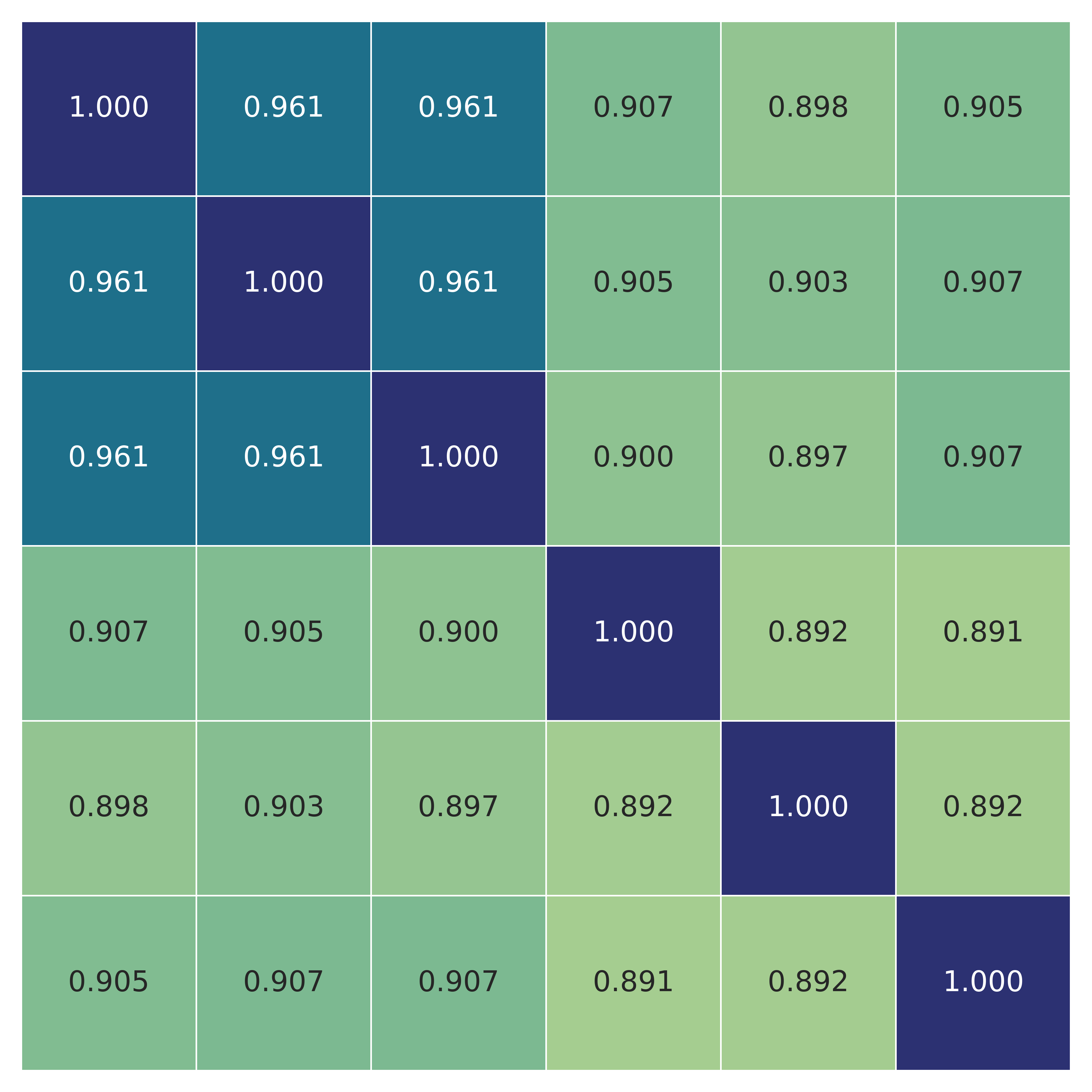}}
\caption{$\kappa=0.2, \tau=1$ (left) and $\tau=20$ (right).}
\label{cross-cka-extra}
\end{center}
\vskip -0.2in
\end{figure}

\begin{figure}[ht]
\vskip 0.2in
\begin{center}
{\includegraphics[width=0.3\columnwidth]{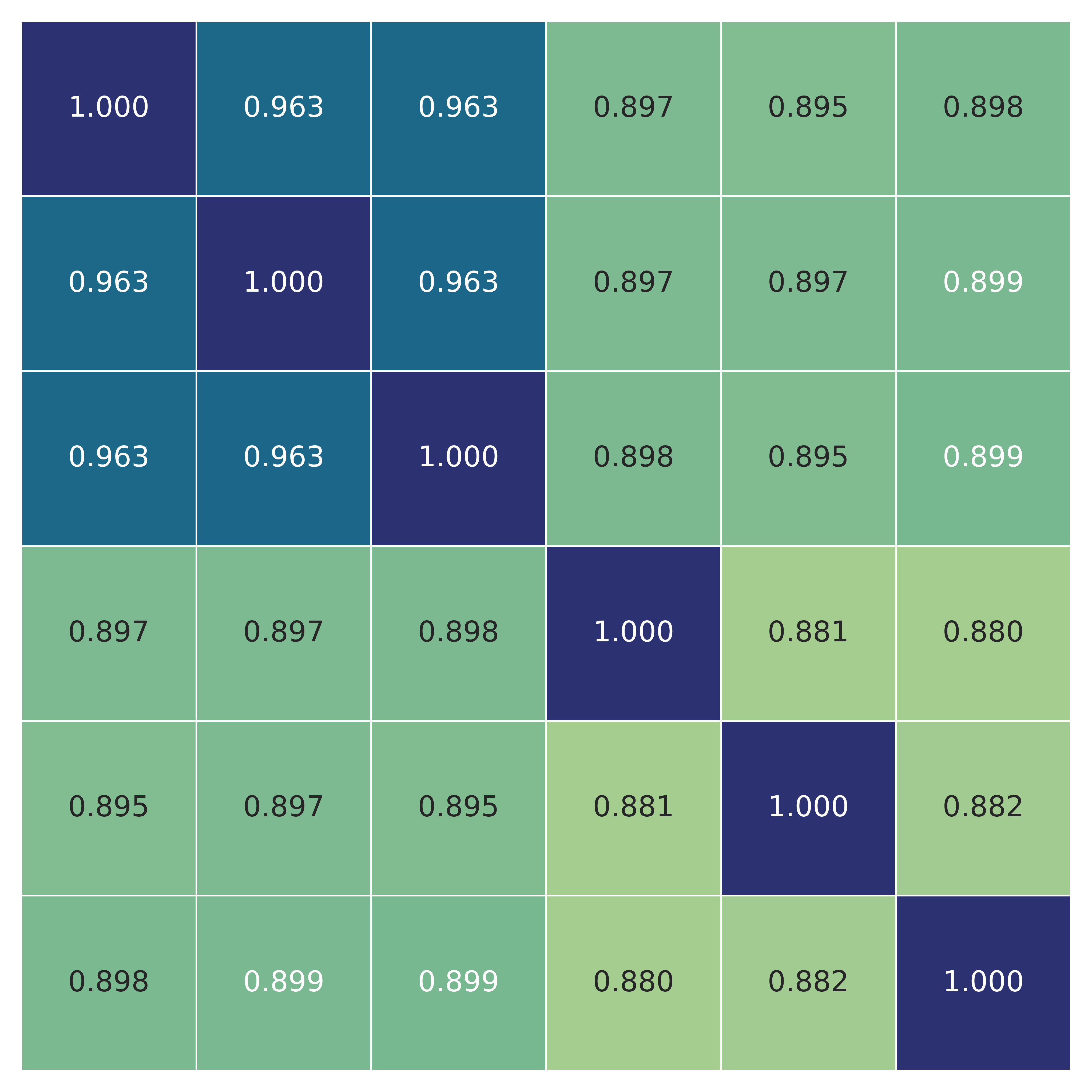}}
{\includegraphics[width=0.3\columnwidth]{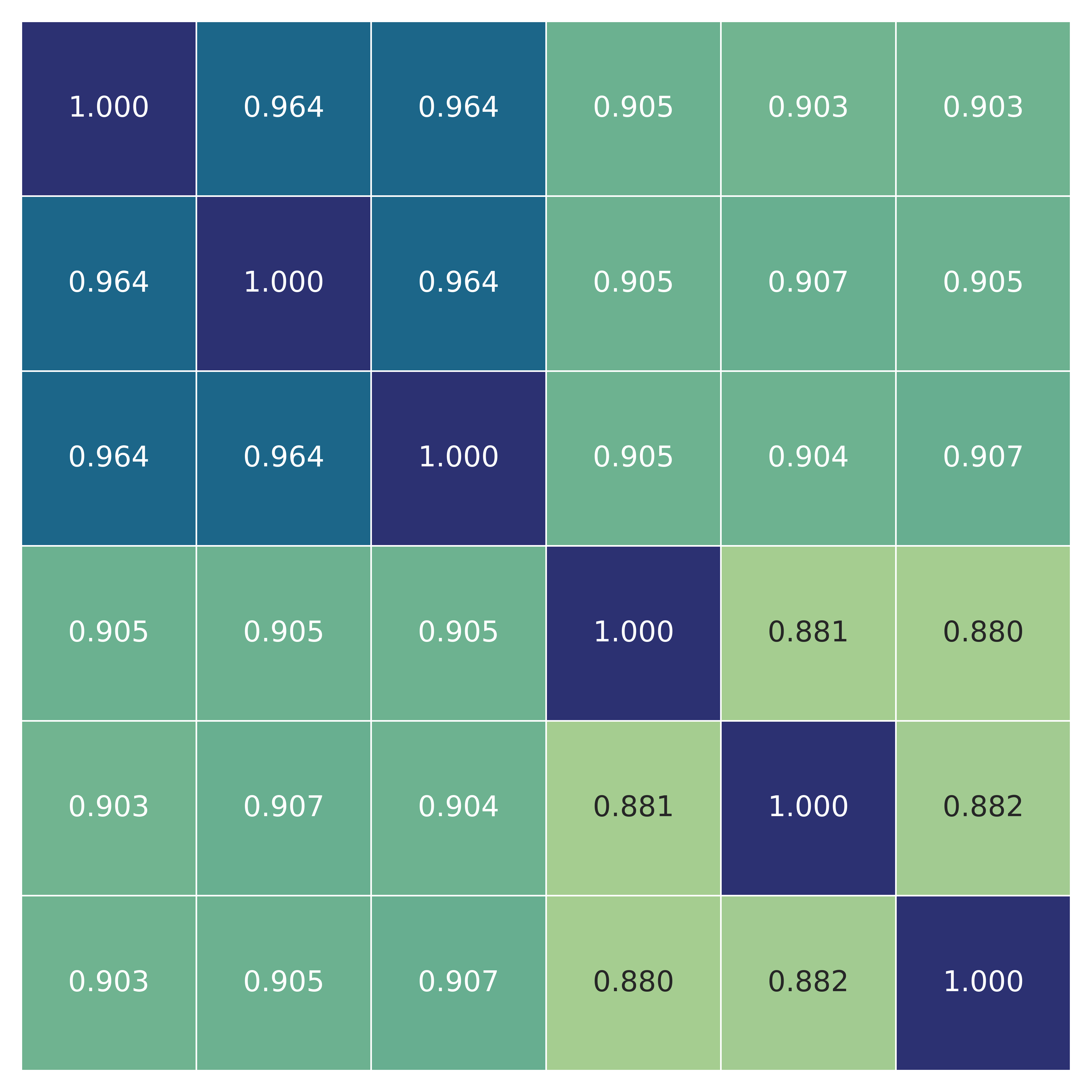}}
\caption{$\kappa=0.4, \tau=1$ (left) and $\tau=20$ (right).}
\label{cross-cka-extra}
\end{center}
\vskip -0.2in
\end{figure}

\begin{figure}[ht]
\vskip 0.2in
\begin{center}
{\includegraphics[width=0.3\columnwidth]{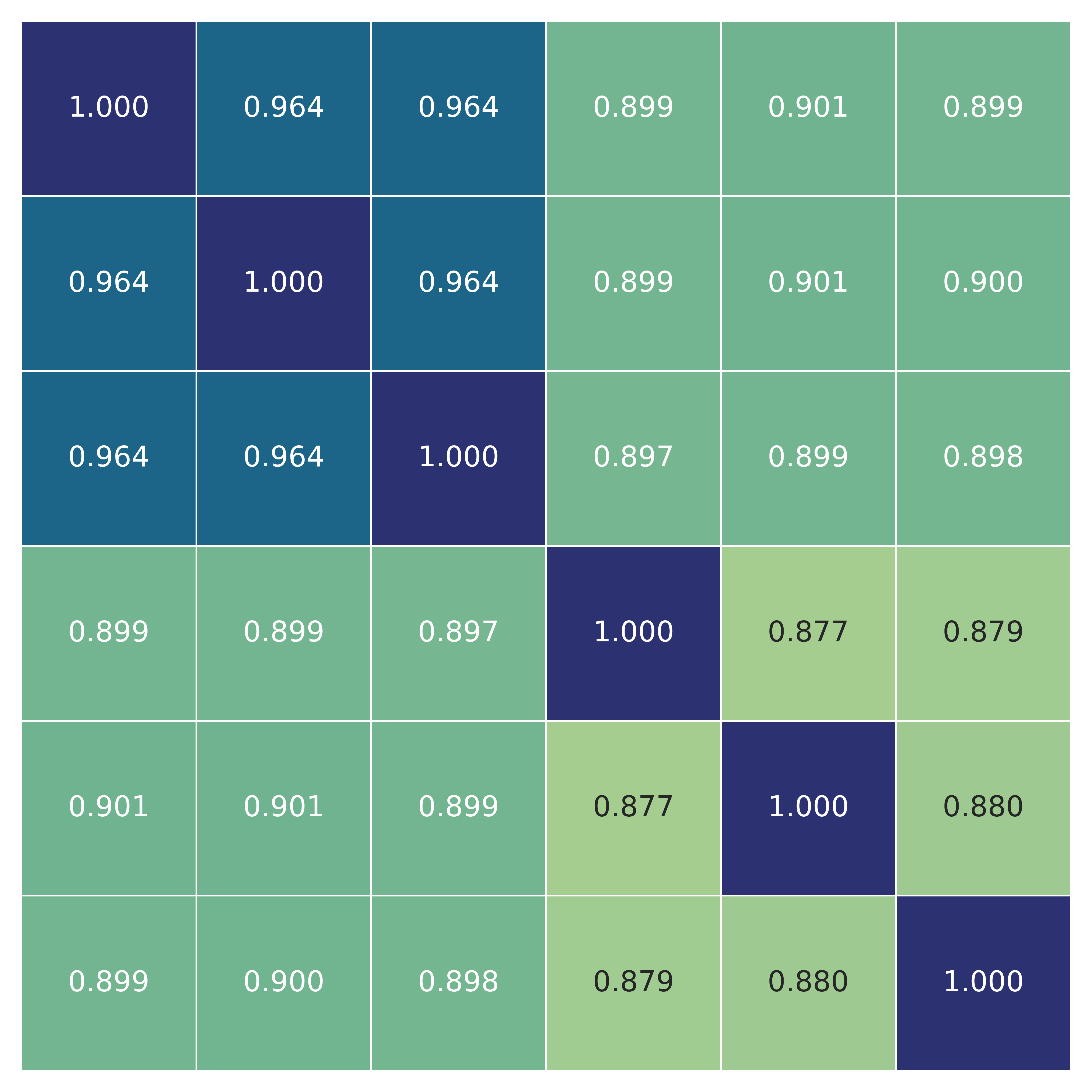}}
{\includegraphics[width=0.3\columnwidth]{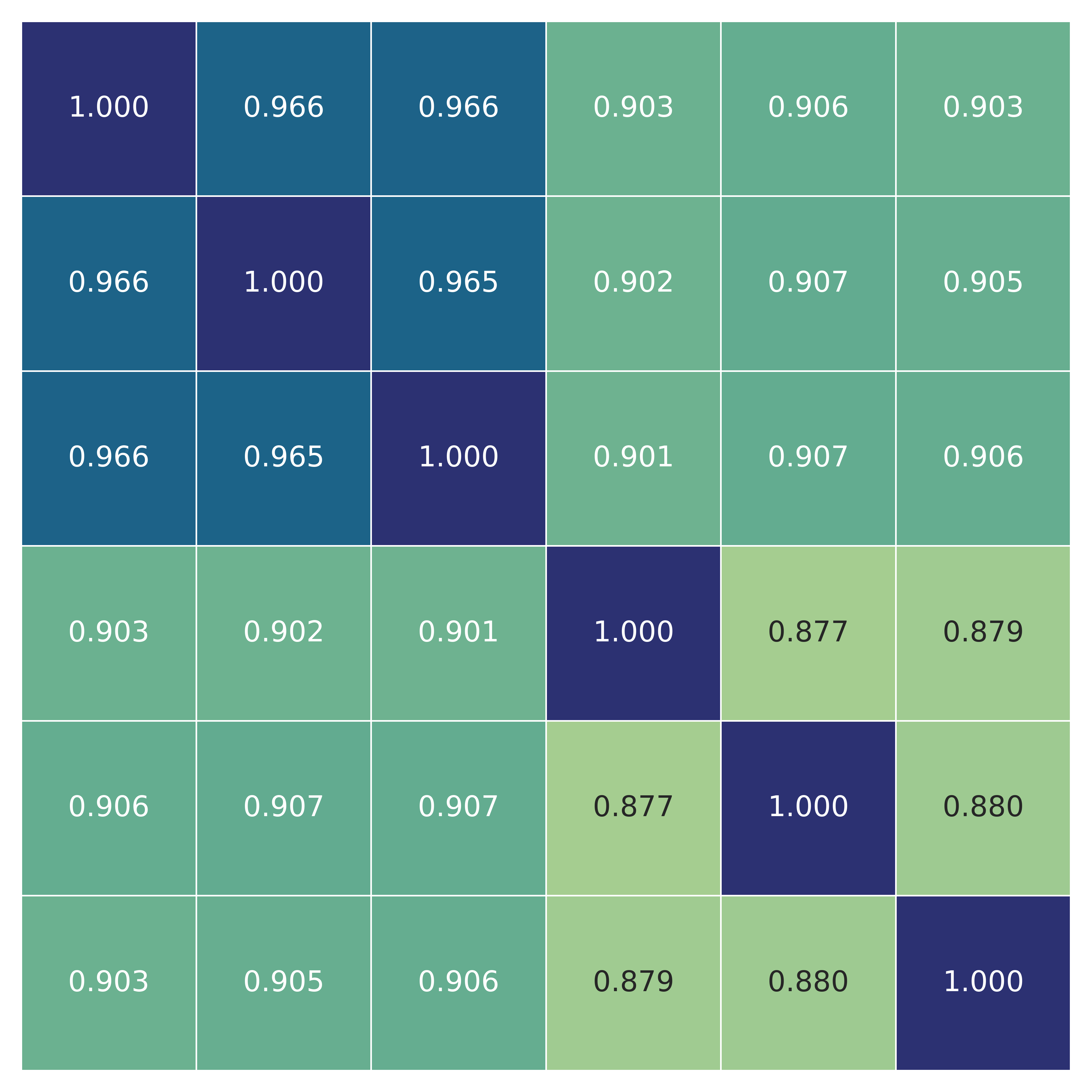}}
\caption{$\kappa=1.0, \tau=1$ (left) and $\tau=20$ (right).}
\label{cross-cka-extra}
\end{center}
\vskip -0.2in
\end{figure}

\end{document}

%% file: defs.tex
\usepackage[utf8]{inputenc}
\usepackage{amsmath,amsfonts}
\usepackage{MnSymbol,stmaryrd,soul}
\usepackage{graphicx}
\usepackage{hyperref}
\usepackage{MnSymbol,stmaryrd}
\usepackage{soul} 
\usepackage{mdframed}
\usepackage{bbm}
\usepackage{minitoc}

\newcommand{\E}{\mathbb E}

\definecolor{textbox1}{RGB}{247, 230, 203}


%% file: math_commands.tex
\usepackage{amsmath,amsfonts,bm}

\def\eqref#1{equation~\ref{#1}}

\def\1{\bm{1}}

\DeclareMathAlphabet{\mathsfit}{\encodingdefault}{\sfdefault}{m}{sl}
\SetMathAlphabet{\mathsfit}{bold}{\encodingdefault}{\sfdefault}{bx}{n}















%% file: arxiv.bbl
\begin{thebibliography}{57}
\providecommand{\natexlab}[1]{#1}
\providecommand{\url}[1]{\texttt{#1}}
\expandafter\ifx\csname urlstyle\endcsname\relax
  \providecommand{\doi}[1]{doi: #1}\else
  \providecommand{\doi}{doi: \begingroup \urlstyle{rm}\Url}\fi

\bibitem[Alain \& Bengio(2016)Alain and Bengio]{alain2016understanding}
Alain, G. and Bengio, Y.
\newblock Understanding intermediate layers using linear classifier probes.
\newblock \emph{arXiv preprint arXiv:1610.01644}, 2016.

\bibitem[Allen-Zhu \& Li(2020)Allen-Zhu and Li]{allen2020towards}
Allen-Zhu, Z. and Li, Y.
\newblock Towards understanding ensemble, knowledge distillation and self-distillation in deep learning.
\newblock \emph{arXiv preprint arXiv:2012.09816}, 2020.

\bibitem[Beyer et~al.(2022)Beyer, Zhai, Royer, Markeeva, Anil, and Kolesnikov]{beyer2022knowledge}
Beyer, L., Zhai, X., Royer, A., Markeeva, L., Anil, R., and Kolesnikov, A.
\newblock Knowledge distillation: A good teacher is patient and consistent.
\newblock In \emph{Proceedings of the IEEE/CVF conference on computer vision and pattern recognition}, pp.\  10925--10934, 2022.

\bibitem[Bordelon et~al.(2020)Bordelon, Canatar, and Pehlevan]{bordelon2020spectrum}
Bordelon, B., Canatar, A., and Pehlevan, C.
\newblock Spectrum dependent learning curves in kernel regression and wide neural networks.
\newblock In \emph{International Conference on Machine Learning}, pp.\  1024--1034. PMLR, 2020.

\bibitem[Buciluǎ et~al.(2006)Buciluǎ, Caruana, and Niculescu-Mizil]{bucilua2006model}
Buciluǎ, C., Caruana, R., and Niculescu-Mizil, A.
\newblock Model compression.
\newblock In \emph{Proceedings of the 12th ACM SIGKDD international conference on Knowledge discovery and data mining}, pp.\  535--541, 2006.

\bibitem[Caron et~al.(2021)Caron, Touvron, Misra, J{\'e}gou, Mairal, Bojanowski, and Joulin]{caron2021emerging}
Caron, M., Touvron, H., Misra, I., J{\'e}gou, H., Mairal, J., Bojanowski, P., and Joulin, A.
\newblock Emerging properties in self-supervised vision transformers.
\newblock In \emph{Proceedings of the IEEE/CVF international conference on computer vision}, 2021.

\bibitem[Chizat et~al.(2019)Chizat, Oyallon, and Bach]{chizat2019lazy}
Chizat, L., Oyallon, E., and Bach, F.
\newblock On lazy training in differentiable programming.
\newblock \emph{Advances in neural information processing systems}, 32, 2019.

\bibitem[Cortes et~al.(2012)Cortes, Mohri, and Rostamizadeh]{cortes2012algorithms}
Cortes, C., Mohri, M., and Rostamizadeh, A.
\newblock Algorithms for learning kernels based on centered alignment.
\newblock \emph{The Journal of Machine Learning Research}, 13\penalty0 (1):\penalty0 795--828, 2012.

\bibitem[Deng et~al.(2009)Deng, Dong, Socher, Li, Li, and Fei-Fei]{deng2009imagenet}
Deng, J., Dong, W., Socher, R., Li, L.-J., Li, K., and Fei-Fei, L.
\newblock Imagenet: A large-scale hierarchical image database.
\newblock In \emph{2009 IEEE Conference on Computer Vision and Pattern Recognition}, pp.\  248--255, 2009.
\newblock \doi{10.1109/CVPR.2009.5206848}.

\bibitem[Dong et~al.(2019)Dong, Hou, Lu, and Zhang]{dong2019distillation}
Dong, B., Hou, J., Lu, Y., and Zhang, Z.
\newblock Distillation $\approx$ early stopping? harvesting dark knowledge utilizing anisotropic information retrieval for overparameterized neural network.
\newblock \emph{arXiv preprint arXiv:1910.01255}, 2019.

\bibitem[Dosovitskiy et~al.(2020)Dosovitskiy, Beyer, Kolesnikov, Weissenborn, Zhai, Unterthiner, Dehghani, Minderer, Heigold, Gelly, et~al.]{dosovitskiy2020image}
Dosovitskiy, A., Beyer, L., Kolesnikov, A., Weissenborn, D., Zhai, X., Unterthiner, T., Dehghani, M., Minderer, M., Heigold, G., Gelly, S., et~al.
\newblock An image is worth 16x16 words: Transformers for image recognition at scale.
\newblock \emph{arXiv preprint arXiv:2010.11929}, 2020.

\bibitem[Foster et~al.(2019)Foster, Greenberg, Kale, Luo, Mohri, and Sridharan]{foster2019hypothesis}
Foster, D.~J., Greenberg, S., Kale, S., Luo, H., Mohri, M., and Sridharan, K.
\newblock Hypothesis set stability and generalization.
\newblock \emph{Advances in Neural Information Processing Systems}, 32, 2019.

\bibitem[Furlanello et~al.(2018)Furlanello, Lipton, Tschannen, Itti, and Anandkumar]{furlanello2018born}
Furlanello, T., Lipton, Z., Tschannen, M., Itti, L., and Anandkumar, A.
\newblock Born again neural networks.
\newblock In \emph{International Conference on Machine Learning}, pp.\  1607--1616. PMLR, 2018.

\bibitem[He \& Ozay(2021)He and Ozay]{he2021feature}
He, B. and Ozay, M.
\newblock Feature kernel distillation.
\newblock In \emph{International Conference on Learning Representations}, 2021.

\bibitem[He \& Ozay(2022)He and Ozay]{pmlr-v162-he22c}
He, B. and Ozay, M.
\newblock Exploring the gap between collapsed \&; whitened features in self-supervised learning.
\newblock In Chaudhuri, K., Jegelka, S., Song, L., Szepesvari, C., Niu, G., and Sabato, S. (eds.), \emph{Proceedings of the 39th International Conference on Machine Learning}, volume 162 of \emph{Proceedings of Machine Learning Research}, pp.\  8613--8634. PMLR, 17--23 Jul 2022.
\newblock URL \url{https://proceedings.mlr.press/v162/he22c.html}.

\bibitem[He et~al.()He, Ding, Guo, Gong, Qin, and Liu]{heDAKDDifficultyAwareKnowledge}
He, C., Ding, Y., Guo, J., Gong, R., Qin, H., and Liu, X.
\newblock {{DA-KD}}: {{Difficulty-Aware Knowledge Distillation}} for {{Efficient Large Language Models}}.

\bibitem[He et~al.(2016)He, Zhang, Ren, and Sun]{he2016deep}
He, K., Zhang, X., Ren, S., and Sun, J.
\newblock Deep residual learning for image recognition.
\newblock In \emph{Proceedings of the IEEE conference on computer vision and pattern recognition}, pp.\  770--778, 2016.

\bibitem[Hinton et~al.(2015)Hinton, Vinyals, and Dean]{hinton2015distilling}
Hinton, G., Vinyals, O., and Dean, J.
\newblock Distilling the knowledge in a neural network.
\newblock \emph{arXiv preprint arXiv:1503.02531}, 2015.

\bibitem[Hsieh et~al.(2023)Hsieh, Li, Yeh, Nakhost, Fujii, Ratner, Krishna, Lee, and Pfister]{hsieh2023distilling}
Hsieh, C.-Y., Li, C.-L., Yeh, C.-K., Nakhost, H., Fujii, Y., Ratner, A., Krishna, R., Lee, C.-Y., and Pfister, T.
\newblock Distilling step-by-step! outperforming larger language models with less training data and smaller model sizes.
\newblock \emph{arXiv preprint arXiv:2305.02301}, 2023.

\bibitem[Jacot et~al.(2018)Jacot, Gabriel, and Hongler]{jacot2018neural}
Jacot, A., Gabriel, F., and Hongler, C.
\newblock Neural tangent kernel: Convergence and generalization in neural networks.
\newblock \emph{Advances in neural information processing systems}, 31, 2018.

\bibitem[Ji \& Zhu(2020)Ji and Zhu]{ji2020knowledge}
Ji, G. and Zhu, Z.
\newblock Knowledge distillation in wide neural networks: Risk bound, data efficiency and imperfect teacher.
\newblock \emph{Advances in Neural Information Processing Systems}, 33:\penalty0 20823--20833, 2020.

\bibitem[Kim \& Kim(2024)Kim and Kim]{kim2024fixed}
Kim, H. and Kim, K.
\newblock Fixed non-negative orthogonal classifier: Inducing zero-mean neural collapse with feature dimension separation.
\newblock In \emph{The Twelfth International Conference on Learning Representations}, 2024.
\newblock URL \url{https://openreview.net/forum?id=F4bmOrmUwc}.

\bibitem[Kingma \& Ba(2017)Kingma and Ba]{kingma2017adam}
Kingma, D.~P. and Ba, J.
\newblock Adam: A method for stochastic optimization.
\newblock \emph{arXiv preprint 1412.6980}, 2017.

\bibitem[Kornblith et~al.(2019)Kornblith, Norouzi, Lee, and Hinton]{kornblith2019similarity}
Kornblith, S., Norouzi, M., Lee, H., and Hinton, G.
\newblock Similarity of neural network representations revisited.
\newblock In \emph{International conference on machine learning}, pp.\  3519--3529. PMLR, 2019.

\bibitem[Krizhevsky \& Hinton(2009)Krizhevsky and Hinton]{krizhevsky2009cifar}
Krizhevsky, A. and Hinton, G.
\newblock Learning multiple layers of features from tiny images.
\newblock Technical report, University of Toronto, 2009.
\newblock URL \url{https://www.cs.toronto.edu/~kriz/learning-features-2009-TR.pdf}.

\bibitem[Liu et~al.(2025)Liu, Yin, and Wang]{liuTheoreticalPerspectivesKnowledge2025a}
Liu, C., Yin, H., and Wang, X.
\newblock Theoretical {{Perspectives}} on {{Knowledge Distillation}}: {{A Review}}.
\newblock \emph{WIREs Computational Statistics}, 17\penalty0 (4):\penalty0 e70049, 2025.
\newblock ISSN 1939-0068.
\newblock \doi{10.1002/wics.70049}.

\bibitem[Lopez-Paz et~al.(2015)Lopez-Paz, Bottou, Sch{\"o}lkopf, and Vapnik]{lopez2015unifying}
Lopez-Paz, D., Bottou, L., Sch{\"o}lkopf, B., and Vapnik, V.
\newblock Unifying distillation and privileged information.
\newblock \emph{arXiv preprint arXiv:1511.03643}, 2015.

\bibitem[Menon et~al.(2021)Menon, Rawat, Reddi, Kim, and Kumar]{menon2021statistical}
Menon, A.~K., Rawat, A.~S., Reddi, S., Kim, S., and Kumar, S.
\newblock A statistical perspective on distillation.
\newblock In \emph{International Conference on Machine Learning}, pp.\  7632--7642. PMLR, 2021.

\bibitem[Mirzadeh et~al.(2020)Mirzadeh, Farajtabar, Li, Levine, Matsukawa, and Ghasemzadeh]{mirzadeh2020improved}
Mirzadeh, S.~I., Farajtabar, M., Li, A., Levine, N., Matsukawa, A., and Ghasemzadeh, H.
\newblock Improved knowledge distillation via teacher assistant.
\newblock In \emph{Proceedings of the AAAI conference on artificial intelligence}, volume~34, pp.\  5191--5198, 2020.

\bibitem[Mobahi et~al.(2020)Mobahi, Farajtabar, and Bartlett]{mobahi2020self}
Mobahi, H., Farajtabar, M., and Bartlett, P.
\newblock Self-distillation amplifies regularization in hilbert space.
\newblock \emph{Advances in Neural Information Processing Systems}, 33:\penalty0 3351--3361, 2020.

\bibitem[Moslemi et~al.(2024)Moslemi, Briskina, Dang, and Li]{moslemiSurveyKnowledgeDistillation2024a}
Moslemi, A., Briskina, A., Dang, Z., and Li, J.
\newblock A survey on knowledge distillation: {{Recent}} advancements.
\newblock \emph{Machine Learning with Applications}, 18:\penalty0 100605, December 2024.
\newblock ISSN 2666-8270.
\newblock \doi{10.1016/j.mlwa.2024.100605}.

\bibitem[M{\"u}ller et~al.(2019)M{\"u}ller, Kornblith, and Hinton]{muller2019does}
M{\"u}ller, R., Kornblith, S., and Hinton, G.~E.
\newblock When does label smoothing help?
\newblock \emph{Advances in neural information processing systems}, 32, 2019.

\bibitem[Nagarajan et~al.(2023)Nagarajan, Menon, Bhojanapalli, Mobahi, and Kumar]{nagarajan2023student}
Nagarajan, V., Menon, A.~K., Bhojanapalli, S., Mobahi, H., and Kumar, S.
\newblock On student-teacher deviations in distillation: does it pay to disobey?
\newblock \emph{arXiv preprint arXiv:2301.12923}, 2023.

\bibitem[Nakkiran et~al.(2020)Nakkiran, Neyshabur, and Sedghi]{nakkiran2020cifar5m}
Nakkiran, P., Neyshabur, B., and Sedghi, H.
\newblock The deep bootstrap framework: Good online learners are good offline generalizers.
\newblock \emph{arXiv preprint arXiv:2010.08127}, 2020.

\bibitem[Panahi et~al.(2022)Panahi, Rahbar, Bhattacharyya, Dubhashi, and Haghir~Chehreghani]{panahi2022analysis}
Panahi, A., Rahbar, A., Bhattacharyya, C., Dubhashi, D., and Haghir~Chehreghani, M.
\newblock Analysis of knowledge transfer in kernel regime.
\newblock In \emph{Proceedings of the 31st ACM International Conference on Information \& Knowledge Management}, pp.\  1615--1624, 2022.

\bibitem[Papyan et~al.(2020)Papyan, Han, and Donoho]{papyan2020prevalence}
Papyan, V., Han, X., and Donoho, D.~L.
\newblock Prevalence of neural collapse during the terminal phase of deep learning training.
\newblock \emph{Proceedings of the National Academy of Sciences}, 117\penalty0 (40):\penalty0 24652--24663, 2020.

\bibitem[Park et~al.(2019)Park, Kim, Lu, and Cho]{park2019relational}
Park, W., Kim, D., Lu, Y., and Cho, M.
\newblock Relational knowledge distillation.
\newblock In \emph{Proceedings of the IEEE/CVF Conference on Computer Vision and Pattern Recognition}, pp.\  3967--3976, 2019.

\bibitem[Passalis \& Tefas(2018)Passalis and Tefas]{passalis2018learning}
Passalis, N. and Tefas, A.
\newblock Learning deep representations with probabilistic knowledge transfer.
\newblock In \emph{Proceedings of the European Conference on Computer Vision (ECCV)}, pp.\  268--284, 2018.

\bibitem[Phuong \& Lampert(2019)Phuong and Lampert]{phuong2019towards}
Phuong, M. and Lampert, C.
\newblock Towards understanding knowledge distillation.
\newblock In \emph{International conference on machine learning}, pp.\  5142--5151. PMLR, 2019.

\bibitem[Sarnthein et~al.(2023)Sarnthein, Bachmann, Anagnostidis, and Hofmann]{sarnthein2023random}
Sarnthein, F., Bachmann, G., Anagnostidis, S., and Hofmann, T.
\newblock Random teachers are good teachers.
\newblock In \emph{Proceedings of the 40th International Conference on Machine Learning}, 2023.

\bibitem[Stanić et~al.(2023)Stanić, Ashley, Serikov, Kirsch, Faccio, Schmidhuber, Hofmann, and Schlag]{languinikitchen}
Stanić, A., Ashley, D., Serikov, O., Kirsch, L., Faccio, F., Schmidhuber, J., Hofmann, T., and Schlag, I.
\newblock The languini kitchen: Enabling language modelling research at different scales of compute.
\newblock \emph{arXiv preprint arXiv:2309.11197}, 2023.

\bibitem[Stanton et~al.(2021)Stanton, Izmailov, Kirichenko, Alemi, and Wilson]{stanton2021does}
Stanton, S., Izmailov, P., Kirichenko, P., Alemi, A.~A., and Wilson, A.~G.
\newblock Does knowledge distillation really work?
\newblock \emph{Advances in Neural Information Processing Systems}, 34:\penalty0 6906--6919, 2021.

\bibitem[Szegedy et~al.(2016)Szegedy, Vanhoucke, Ioffe, Shlens, and Wojna]{szegedy2016rethinking}
Szegedy, C., Vanhoucke, V., Ioffe, S., Shlens, J., and Wojna, Z.
\newblock Rethinking the inception architecture for computer vision.
\newblock In \emph{Proceedings of the IEEE conference on computer vision and pattern recognition}, pp.\  2818--2826, 2016.

\bibitem[Tang et~al.(2020)Tang, Shivanna, Zhao, Lin, Singh, Chi, and Jain]{tang2020understanding}
Tang, J., Shivanna, R., Zhao, Z., Lin, D., Singh, A., Chi, E.~H., and Jain, S.
\newblock Understanding and improving knowledge distillation.
\newblock \emph{arXiv preprint arXiv:2002.03532}, 2020.

\bibitem[Tian et~al.(2020)Tian, Krishnan, and Isola]{Tian2020Contrastive}
Tian, Y., Krishnan, D., and Isola, P.
\newblock Contrastive representation distillation.
\newblock In \emph{International Conference on Learning Representations}, 2020.
\newblock URL \url{https://openreview.net/forum?id=SkgpBJrtvS}.

\bibitem[Touvron et~al.(2021)Touvron, Cord, Douze, Massa, Sablayrolles, and Jegou]{touvron2021deit}
Touvron, H., Cord, M., Douze, M., Massa, F., Sablayrolles, A., and Jegou, H.
\newblock Training data-efficient image transformers \& distillation through attention.
\newblock In \emph{Proceedings of the 38th International Conference on Machine Learning}, 2021.

\bibitem[Tung \& Mori(2019)Tung and Mori]{tung2019similarity}
Tung, F. and Mori, G.
\newblock Similarity-preserving knowledge distillation.
\newblock In \emph{Proceedings of the IEEE/CVF International Conference on Computer Vision}, pp.\  1365--1374, 2019.

\bibitem[Wu et~al.(2025)Wu, Jiang, Li, Zhai, Liu, Hao, Liu, Yang, Xie, Gu, Yang, Zhang, Bao, and Wang]{wuScalingLawDataEfficient2025}
Wu, X., Jiang, X., Li, H., Zhai, J., Liu, D., Hao, Q., Liu, H., Yang, Z., Xie, J., Gu, N., Yang, J., Zhang, K., Bao, Y., and Wang, J.
\newblock Beyond {{Scaling Law}}: {{A Data-Efficient Distillation Framework}} for {{Reasoning}}, August 2025.

\bibitem[Xu et~al.()Xu, Wilf, Liang, Obolenskiy, Fried, and Morency]{xuComparativeKnowledgeDistillationa}
Xu, A.~T., Wilf, A., Liang, P.~P., Obolenskiy, A., Fried, D., and Morency, L.-P.
\newblock Comparative {{Knowledge Distillation}}.

\bibitem[Yang \& Hu(2020)Yang and Hu]{yang2020feature}
Yang, G. and Hu, E.~J.
\newblock Feature learning in infinite-width neural networks.
\newblock \emph{arXiv preprint arXiv:2011.14522}, 2020.

\bibitem[Yim et~al.(2017)Yim, Joo, Bae, and Kim]{yim2017gift}
Yim, J., Joo, D., Bae, J., and Kim, J.
\newblock A gift from knowledge distillation: Fast optimization, network minimization and transfer learning.
\newblock In \emph{Proceedings of the IEEE conference on computer vision and pattern recognition}, pp.\  4133--4141, 2017.

\bibitem[Yuan et~al.(2020)Yuan, Tay, Li, Wang, and Feng]{yuan2020revisiting}
Yuan, L., Tay, F.~E., Li, G., Wang, T., and Feng, J.
\newblock Revisiting knowledge distillation via label smoothing regularization.
\newblock In \emph{Proceedings of the IEEE/CVF Conference on Computer Vision and Pattern Recognition}, pp.\  3903--3911, 2020.

\bibitem[Zagoruyko \& Komodakis(2016)Zagoruyko and Komodakis]{zagoruyko2016paying}
Zagoruyko, S. and Komodakis, N.
\newblock Paying more attention to attention: Improving the performance of convolutional neural networks via attention transfer.
\newblock \emph{arXiv preprint arXiv:1612.03928}, 2016.

\bibitem[Zhang et~al.(2023)Zhang, Lyu, and Chen]{zhang2023revisiting}
Zhang, S., Lyu, Z., and Chen, X.
\newblock Revisiting knowledge distillation under distribution shift.
\newblock \emph{arXiv preprint arXiv:2312.16242}, 2023.

\bibitem[Zhao et~al.(2022)Zhao, Cui, Song, Qiu, and Liang]{zhao2022decoupled}
Zhao, B., Cui, Q., Song, R., Qiu, Y., and Liang, J.
\newblock Decoupled knowledge distillation.
\newblock In \emph{Proceedings of the IEEE/CVF Conference on computer vision and pattern recognition}, pp.\  11953--11962, 2022.

\bibitem[Zhao \& Zhu(2023)Zhao and Zhu]{zhao2023towards}
Zhao, Q. and Zhu, B.
\newblock Towards the fundamental limits of knowledge transfer over finite domains.
\newblock \emph{arXiv preprint arXiv:2310.07838}, 2023.

\bibitem[Zhou et~al.(2021)Zhou, Song, Chen, Zhou, Wang, Yuan, and Zhang]{zhou2021rethinking}
Zhou, H., Song, L., Chen, J., Zhou, Y., Wang, G., Yuan, J., and Zhang, Q.
\newblock Rethinking soft labels for knowledge distillation: A bias-variance tradeoff perspective.
\newblock \emph{arXiv preprint arXiv:2102.00650}, 2021.

\end{thebibliography}
